\documentclass[runningheads]{llncs}
\usepackage{graphicx}

\usepackage{tikz}
\usepackage{comment}
\usepackage{amsmath,amssymb} 
\usepackage{color}
\usepackage{booktabs}
\usepackage{multirow}
\usepackage{times}
\usepackage{soul}
\usepackage{indentfirst}
\usepackage{breakcites}

\usepackage{enumitem}

\usepackage[accsupp]{axessibility}  
\usepackage[pagebackref=true,breaklinks=true,letterpaper=true,colorlinks,bookmarks=false]{hyperref}

\usepackage{caption}

\usepackage[capitalize]{cleveref}
\crefname{section}{Sec.}{Secs.}
\Crefname{section}{Section}{Sections}
\Crefname{table}{Table}{Tables}
\crefname{table}{Tab.}{Tabs.}

\begin{document}
\pagestyle{headings}
\mainmatter
\def\ECCVSubNumber{2039}  

\title{3D-Aware Semantic-Guided Generative Model for Human Synthesis} 


\titlerunning{3D-SGAN}
%


\author{Jichao Zhang \inst{1} \and
Enver Sangineto \inst{2} \and 
Hao Tang\inst{3} \and
Aliaksandr Siarohin \inst{1,4} \and \\ 
Zhun Zhong \inst{1} \and 
Nicu Sebe \inst{1} \and 
Wei Wang \inst{1}}
\authorrunning{J. Zhang et al.}
%
\institute{$^{1}$University of Trento~
$^{2}$University of Modena and Reggio Emilia~
$^{3}$ETH Zurich~~$^{4}$Snap Research 
}

\maketitle

\begin{abstract}
Generative Neural Radiance Field (GNeRF) models, which extract implicit 3D representations from 2D images, have recently been shown to produce realistic images representing rigid/semi-rigid objects, such as human faces or cars. However, they usually struggle to generate high-quality images representing non-rigid objects, such as the human body, which is of a great interest for many computer graphics applications. This paper proposes a 3D-aware Semantic-Guided Generative Model (3D-SGAN) for human image synthesis, which combines a GNeRF with a texture generator. The former learns an implicit 3D representation of the human body and outputs a set of 2D semantic segmentation masks. The latter transforms these semantic masks into a real image, adding a realistic texture to the human appearance. Without requiring additional 3D information, our model can learn 3D human representations with a photo-realistic, controllable generation. Our experiments on the DeepFashion dataset show that 3D-SGAN significantly outperforms the most recent baselines. The code is available at \url{https://github.com/zhangqianhui/3DSGAN}.
\keywords{Generative Neural Radiance Fields, Human image generation}
\end{abstract}

\section{Introduction}
\label{Introduction}

Recent deep generative models can generate and manipulate high-quality images. For instance, Generative Adversarial Networks (GANs)~\cite{goodfellow2014generative},  have been applied to different tasks, such as image-to-image translation~\cite{zhu2017unpaired,choi2018stargan,huang2018multimodal}, portrait editing~\cite{abdal2020styleflow,shen2020interpreting,tov2021designing,shen2021closedform}, and semantic image synthesis~\cite{park2019semantic}, to mention a few. However, most state-of-the-art GAN models~\cite{gulrajani2017improved,karras2017progressive,karras2018style,karras2020analyzing,Karras2020ada,he2021eigengan,Karras2021} are trained using 2D images only, operate in the 2D domain, and ignore the 3D nature of the world. Thus, they often struggle to disentangle the underlying 3D factors of the represented objects. 

\begin{figure}[!t]
\centering
\includegraphics[width=1.0\linewidth]{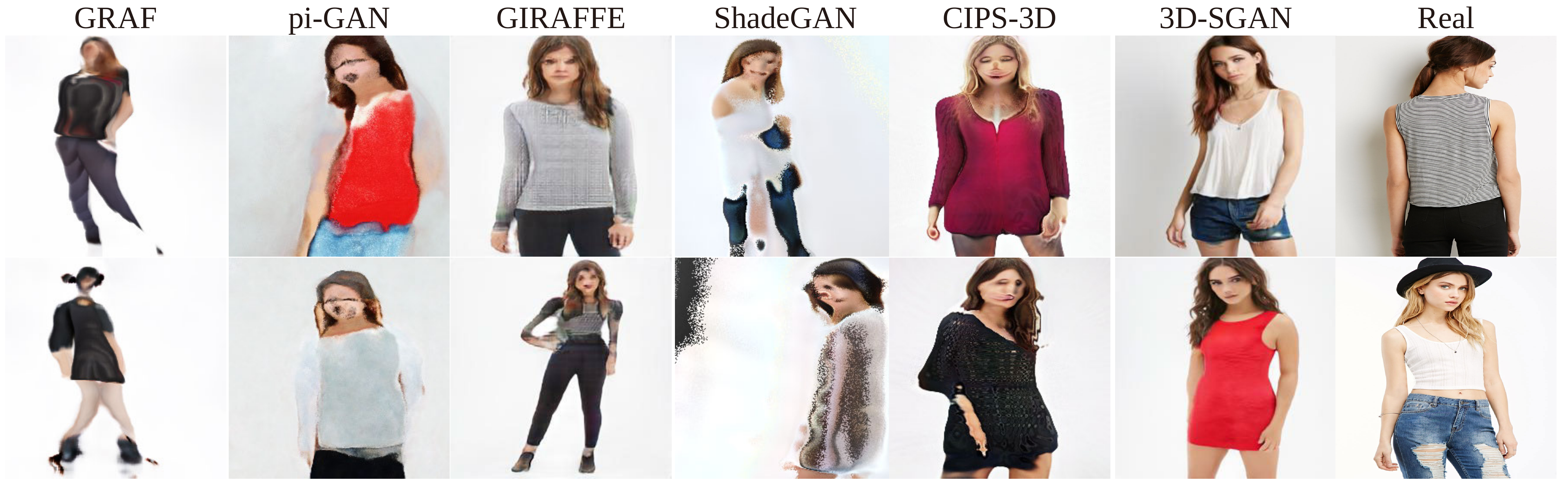}
\caption{A qualitative comparison between different   generation methods: GRAF~\cite{Schwarz2020NEURIPS}, pi-GAN~\cite{chanmonteiro2020pi-GAN}, GIRAFFE~\cite{GIRAFFE}, ShadeGAN~\cite{pan2021shading} CIPS-3D~\cite{zhou2021CIPS3D}, and 3D-SGAN (Ours).}
\label{fig:show}
\end{figure}

Recently, different 3D-aware generative models~\cite{HoloGAN2019,BlockGAN2020,3dgan} have been proposed to solve this problem. Since  most of these methods do not need 3D annotations, they can create 3D content while reducing the  hardware costs of common  computer graphics alternatives. Differently from generating 3D untextured shapes~\cite{3dgan,Gadelha20173DSI}, some of these methods~\cite{VON,Chen2021TowardsAN,HoloGAN2019,Liao2020CVPR,BlockGAN2020} focus on  3D-aware realistic image generation and controllability. Generally speaking, these  models mimic the traditional computer graphics rendering pipeline:  they first model the 3D structure, then they use a (differentiable) projection module to project the 3D structure into 2D images.
The latter may be a depth map~\cite{Chen2021TowardsAN}, a sketch~\cite{VON} or a feature map~\cite{HoloGAN2019} which is finally mapped into the real image by a rendering module. During training, some methods require 3D data~\cite{VON,Chen2021TowardsAN}, and some~\cite{HoloGAN2019,Liao2020CVPR,BlockGAN2020} can learn a 3D representation directly from raw images. 

An important class of {\em implicit} 3D representations are the 
Neural Radiance Fields (NeRFs), which can generate high-quality  unseen views  of complex scenes~\cite{mildenhall2020nerf,jain2021putting,devries2021unconstrained,peng2021neural,peng2021animatable,Reiser2021ICCV,chanmonteiro2020pi-GAN}. Generative NeRFs (GNeRFs) combine NeRFs with GANs in order to condition the generation process with a latent code governing the object's appearance or shape~\cite{Schwarz2020NEURIPS,chanmonteiro2020pi-GAN,GIRAFFE}.
However, these methods~\cite{Schwarz2020NEURIPS,chanmonteiro2020pi-GAN,GIRAFFE,pan2021shading,zhou2021CIPS3D} focus on relatively simple and ``rigid'' objects, such as cars and faces, and they usually struggle to generate highly non-rigid objects such as the human body (e.g., see Fig.~\ref{fig:show}). This is likely due to the fact that the human body appearance is highly variable because of both its articulated poses and the variability of the clothes texture, being these two factors entangled with each other.
Thus, adversarially learning the data distribution modeling all these  factors, is a hard task, especially when the training set is relatively small.

To mitigate this problem, we propose to {\em split} the human generation process in two separate steps and use intermediate segmentation masks as the bridge of these two stages. Specifically, 
our 3D-aware Semantic-Guided Generative model (3D-SGAN) is composed of two generators:  a GNeRF model and a texture generator. The GNeRF model learns the 3D structure of the human body and generates a semantic segmentation of the main body components, which is largely invariant to the surface texture. The texture generator translates the previous segmentation output into a photo-realistic image. 
To control the texture style, a Variational AutoEncoder (VAE~\cite{kingma2013auto}) approach with a StyleGAN-like \cite{karras2020analyzing} decoder
is used to modulate the final decoding process. The similar idea has been used in ~\cite{lassner2017generative}, but their semantic generator is 2D, and it cannot perform 3D manipulations.  
We empirically show that splitting the human generation process into these two stages brings the following three advantages. First, the GNeRF model is able to learn the intrinsic 3D geometry of the human body, even when trained with a  small dataset. Second, the texture generator can successfully translate semantic information into a textured object. Third, both generators can be controlled by explicitly varying their respective conditioning latent codes.
Moreover, we propose two consistency losses to further disentangle  the latent codes representing the garment type (which we call the ``semantic'' code) and the human pose. 
 Finally, since there is no general metric which can be used to evaluate the 3D consistency of image generation with multiple viewpoints, we propose a
 point matching-based metric which we name
 {\em average Matched Points  (aMP)}. Experiments conducted on the  DeepFashion dataset~\cite{liu2016deepfashion} show that  3D-SGAN can generate high-quality person images significantly outperforming state-of-the-art approaches. In summary, the main contributions of this work are:
\begin{itemize}
  \item[1)] We propose 3D-SGAN, which combines a GNeRF with a VAE-conditioned texture generator for human-image synthesis.
  \item[2)] We propose two consistency losses to increase the disentanglement between semantic information (e.g., garment type) and the human pose.
  \item[3)] We show that 3D-SGAN generates high-quality human images, significantly outperforming the previous controllable state-of-the-art methods. 
  \item[4)] We propose a new metric (aMP) to evaluate the 3D-view consistency. 
\end{itemize}

\section{Related work}

\noindent \textbf{3D-aware image synthesis} is based on generative models which incorporate a 3D scene representation. This allows rendering photo-realistic images from different viewpoints. Early methods use GAN-based architectures for building 3D  voxel~\cite{3dgan,Gadelha20173DSI,DBLP:journals/corr/abs-2002-12674,henzler2019platonicgan} or mesh~\cite{10.5555/3157382.3157656,Henderson2019LearningS3} representations. However, they mostly focus on learning untextured 3D structures.
More recently, different methods
learn textured representations directly from 2D images~\cite{Wang2016GenerativeIM,VON,HoloGAN2019,Schwarz2020NEURIPS,devries2021unconstrained,DBLP:journals/corr/abs-2103-17269,GIRAFFE}. The resulting controllable 3D scene representation can be used for image synthesis.
Some of these methods~\cite{VON,Chen2021TowardsAN} require extra 3D data for disentangling shape from texture. The main idea  is to generate an internal 3D shape and then project this shape into 2D sketches~\cite{VON} or depth maps~\cite{Chen2021TowardsAN}, which are finally rendered in a realistic image.
Other methods are directly trained on 2D images without using 3D data~\cite{HoloGAN2019,Schwarz2020NEURIPS,BlockGAN2020,GIRAFFE,xu2021generative,sun2021fenerf}.
For instance,
inspired by StyleGANv2~\cite{karras2020analyzing}, Thu et al.~\cite{HoloGAN2019} propose HoloGAN, which  predicts 3D abstract features using 3D convolutions, and then projects these features into a 2D representation which is finally decoded into an image. However, the learnable projection function, e.g., the decoder, results in an entangled representation, thus the view-consistency of the generated images is degraded.
Katja et  al.~\cite{Schwarz2020NEURIPS}  use a NeRF to represent the 3D scene and a volume rendering technique to render the final image. However, this model works at relatively low image resolutions and it is restricted to single-object scenes.
To tackle these issues, some works propose object-aware scene representations. For example, Liao et al.~\cite{Liao2020CVPR}  combine a 2D generator and a projection module with a 3D generator which outputs multiple abstract 3D primitives. Every stage in this model outputs multiple terms to separately represent each object. Instead of abstract 3D primitives, Phuoc et al.~\cite{BlockGAN2020} use a voxel feature grid as the 3D representation, but their method fails to generate consistent images at high-resolution. Michael et al.~\cite{GIRAFFE} recently introduced GIRAFFE, a multiple-object scene representation based on NeRFs, jointly with
 an object composition operator. GIRAFFE is the state-of-the-art 
3D-aware  approach for both single and multiple object generation tasks.
A few very recent papers~\cite{pan2021shading,tan2022volux} propose to learn an accurate object geometry by introducing a relighting module into the rendering process. 
Xu et al. \cite{xu20213d},  explicitly learn a structural
 and a textural representation (a feature volume), 
which is used jointly with the implicit NeRF mechanism.
Chan et al. 
\cite{chan2021efficient}, propose to replace   
 the 3D volume with three projection feature planes.
Finally, 3D-consistency is addressed in ~\cite{gu2021stylenerf,zhou2021CIPS3D,orel2021stylesdf}, where, e.g.,  StyleGAN-based networks are used for neural rendering. However, most of these works fail to disentangle the semantic attributes. 

While previous methods can achieve impressive rigid-object generation and manipulation results, they usually struggle to deal with non-rigid objects with complex pose and texture variations. 
For instance, the human body is a non-rigid object which is very important in many generative applications.

\begin{figure*}[!t] 
\centering
\includegraphics[width=1.0\linewidth]{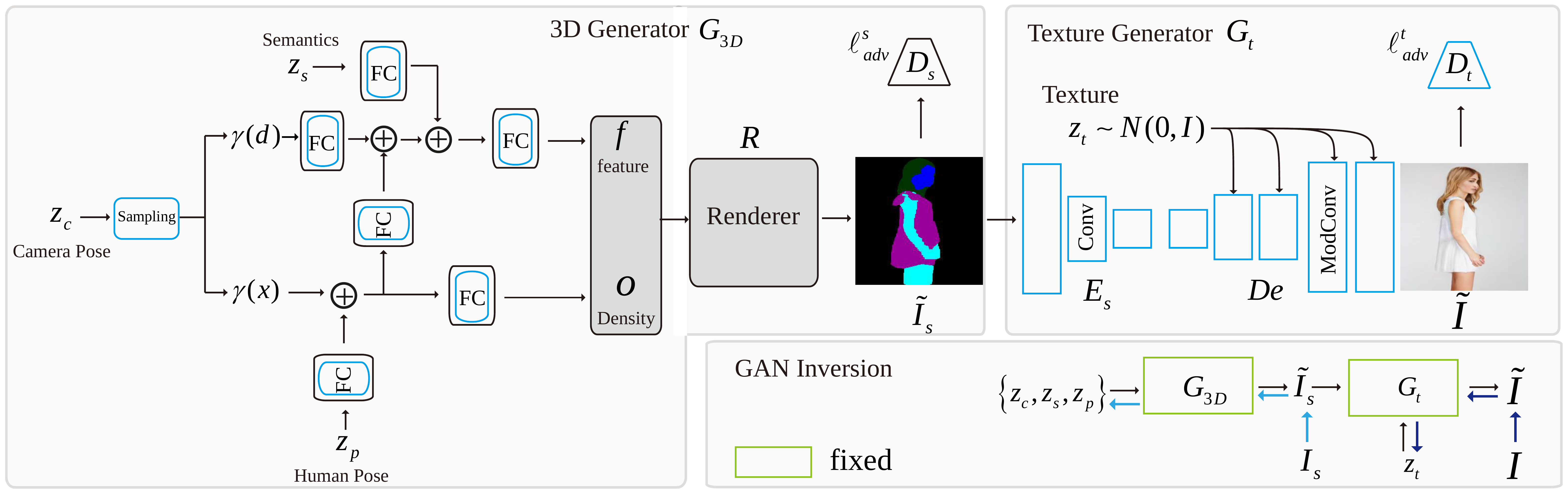}
\caption{An overview of the proposed 3D-SGAN architecture, composed of  two main generators.
$G_{3D}$ (on the left) follows a GNeRF structure, with a NeRF kernel used to represent implicit 3D information, latent codes governing different appearance variations and a discriminator ($D_s$) which is used for  adversarial training. The output of 
$G_{3D}$ is the semantic masks $\tilde I_s$  (middle).
The second   generator ($G_t$, right) translates the semantic masks into a photo-realistic image $\tilde I$.
Also $G_t$ is trained adversarially (see top right, the second discriminator $D_t$).
 The human generation process can be controlled by interpolating different latent codes: the semantics  $\pmb{z}_s$, the pose  $\pmb{z}_p$, the camera  $\pmb{z}_c$, and the texture code $\pmb{z}_t$. The bottom of the figure shows the GAN inversion scheme.}
\label{fig:model}
\end{figure*}

\noindent \textbf{GANs for human generation.}
GANs~\cite{goodfellow2014generative}  have been widely used for different object categories, such as, for instance, faces~\cite{Karras2020AnalyzingAI,karras2017progressive,karras2018style,brock2018large}, cars~\cite{Karras2020AnalyzingAI,karras2018style,Schwarz2020NEURIPS,BlockGAN2020}, and churches~\cite{GIRAFFE}. However,  GANs still struggle to produce high-quality full-human body  images, because of the complex pose variations. Very recently, Sarkar et al.~\cite{sarkar2021humangan} proposed a VAE-GAN model for the pose transfer and the part sampling tasks. In more detail, this model extracts an UV texture map from the input image using DensePose~\cite{alp2018densepose}, and then encodes the texture into a Gaussian distribution. Then, it samples from this distribution and warps the sample into the target pose space. Finally, the warped latent code is used as input to the decoder. Compared to Sarkar et al.~\cite{sarkar2021humangan}, our method does not use an SMPL nor DensePose to extract the point correspondences as additional supervised information. Despite that, 3D-SGAN  can learn 3D representations of the human body and  control the generation process (e.g., by changing   the input camera parameters). 

StylePeople~\cite{Grigorev_2021_CVPR} is based on a full-body human avatar, which combines  StyleGANv2~\cite{karras2020analyzing} with {\em neural dressing}. The StyleGANv2 module samples neural textures, and these textures are superimposed on the meshes of an SMPL. The textured meshes are finally rendered into an image. In contrast, our 3D-SGAN can perform semantic disentanglement and manipulation using semantic codes. 

\noindent \textbf{Pose transfer} aims to synthesize person images in a novel view or in a new pose. This is a very challenging task, since it requires very complicated spatial transformations to account for different poses.  Most works in this field can be categorized  by the way in which the human pose is represented. Early works are based on keypoints~\cite{ma2017pose,siarohin2019deformable,zhu2019progressive,ren2020deep,tang2020xinggan,ma2018disentangled,men2020controllable,huang2020generating,Balakrishnan2018SynthesizingIO,lv2021learning,PISE,siarohin2019first,Zhou_2021_CVPR,Sanyal_2021_ICCV,song2019unsupervised,yildirim2019generating,sanyal2021learning}. More recent methods~\cite{Grigorev2018CoordinatebasedTI,neverova2018dense,Sarkar2020,sarkar2021style,liu2021liquid} use correspondences between pixel location in 2D images and points in SMPL~\cite{SMPL:2015}  (usually estimated using  DensePose~\cite{guler2018densepose}). However, these approaches usually struggle to simultaneously provide a realistic and a 3D controllable person generation.

\section{Preliminaries}
\label{Preliminaries}

NeRF~\cite{mildenhall2020nerf}  is an implicit model which represents  a 3D scene using the weights of a  multilayer perceptron (MLP). This MLP ($h$) takes as input a 3D coordinate $\pmb{x} \in \mathbb{R}^3$ 
 and a view direction $\pmb{d} \in \mathbb{R}^2$, and outputs the  density (or ``opacity", $o$) and the view-dependent RGB color value $\pmb{c}$:
\begin{equation}
\label{nerf}
\begin{aligned}
(\pmb{c}, o) = h(\gamma(\pmb{x}), \gamma(\pmb{d})),
\end{aligned}
\end{equation}
\noindent
where $\gamma$ is a positional encoding function \cite{attention-is-all-you-need}.
 On the other hand,
Generative NeRF (GNeRF)~\cite{Schwarz2020NEURIPS} is a  NeRF conditioned on the latent codes $\pmb{z}_g$ and $\pmb{z}_a$, respectively representing the geometric shape and the object appearance, and drawn from  a priori distributions. GNeRFs~\cite{Schwarz2020NEURIPS,GIRAFFE} are trained using an adversarial approach.
In GIRAFFE~\cite{GIRAFFE}, 
 the color value ($\pmb{c}$ in Eq.~\ref{nerf}) is replaced by an intermediate feature vector $\pmb{f}$:
\begin{equation}
\label{Gnerf}
\begin{aligned}
(\pmb{f}, o) = h(\gamma(\pmb{x}), \gamma(\pmb{d}), \pmb{z}_g, \pmb{z}_a).
\end{aligned}
\end{equation}
\noindent

$\pmb{f}$ is mapped into  a photo-realistic image using a volume and neural rendering module $R$ and fed to a discriminator (more details in \cite{Schwarz2020NEURIPS,GIRAFFE}).

Our 3D generator (see Sec.~\ref{Method}) is inspired by GIRAFFE \cite{GIRAFFE}. However, it learns to produce a segmentation image, a simpler task with respect to directly generating a photo-realistic image (see Sec.~\ref{Introduction}).

\section{The proposed 3D-SGAN}
\label{Method}

Fig.~\ref{fig:model} shows the proposed 3D-SGAN architecture, composed of two main modules: a  3D-based segmentation mask generator and a texture generator. 
The former ($G_{3D}$) generates semantic segmentation masks of the human body which correspond to the main body parts and depend on the type of clothes, the camera viewpoint and the human pose. 
On the other hand, the texture generator ($G_{t}$) takes as input these segmentation masks and translates them into a photo-realistic image, adding a texture style randomly drawn from a pre-learned marginal distribution. 
The two modules are trained separately.

\subsection{3D generator for semantic mask rendering}
\label{G3D}

Given a set of 2D human image samples $\{I^{i}\}^{N}_{i=1}$, we first use an off-the-shelf human parsing tool~\cite{Badrinarayanan2017SegNetAD} to obtain the corresponding ground-truth semantic segmentation masks $\{I^{i}_{s}\}^{N}_{i=1}$. Using  $T = \{(I^{i},I^{i}_{s})\}^{N}_{i=1}$ as our training set, the goal is to train a two-step generative model:
\begin{equation}
\begin{aligned}
\tilde I = G(\pmb{z}_{c}, \pmb{z}_{s}, \pmb{z}_{p}, \pmb{z}_{t})=  G_{t}(
G_{3D}(\pmb{z}_{c}, \pmb{z}_{s}, \pmb{z}_{p}), \pmb{z}_{t}),
\end{aligned}
\end{equation}
where $\tilde I$ is the final  generated image. The latent codes $\pmb{z}_{c} \sim P_{c}$ (see Sec.\ref{exper}), $\pmb{z}_{s} \sim \mathcal{N}(0,\pmb{I})$, $\pmb{z}_{p} \sim \mathcal{N}(0,\pmb{I})$, and $\pmb{z}_{t} \sim \mathcal{N}(0,\pmb{I}) $ represent, respectively: the camera viewpoint, the semantics (i.e., the garment type), the  body pose and the human texture.

The structure of our 3D Generator $G_{3D}$ is inspired by GIRAFFE~\cite{GIRAFFE} (Sec.~\ref{Preliminaries}).
However, differently  from \cite{Schwarz2020NEURIPS,GIRAFFE}, which 
learn to generate a textured object, in our case, $h$ learns to generate a semantically segmented image.
Specifically, we use a latent semantic code ($\pmb{z}_{s}$) to condition the final segmentation output on the type of garment. 
As shown in Fig.~\ref{fig:model}, $\pmb{z}_{s}$ does not influence the opacity generation branch, and it is injected into the direction-dependent branch, which finally outputs a feature vector $\pmb{f}$, representing a point-wise semantic content.
Formally, we have:
\begin{equation}
\begin{aligned}
(\pmb{f}, o) = h(\gamma(\pmb{x}), \gamma(\pmb{d}), \pmb{z}_{c}, \pmb{z}_{s}, \pmb{z}_{p}).
\end{aligned}
\end{equation}
Following \cite{Schwarz2020NEURIPS,GIRAFFE}, we  generate a set of pairs  $\{ (\pmb{f}, o) \}$ which are finally 
projected into the 2D plane using 
 a  rendering module $R$~\cite{mildenhall2020nerf,GIRAFFE}
 (see \cref{Preliminaries}),
and represented by the segmentation masks $\tilde I_s$. 
 Specifically, $\tilde I_s$ is a tensor composed of $n_s$ channels, where each channel represents a segmentation mask of the same spatial resolution of the real images in $T$ (Fig.~\ref{fig:model}).
 
$G_{3D}$  is trained jointly with a discriminator $D_s$, which learns to discriminate between real ($I_s$) and fake ($\tilde I_s$) segmentation masks (more details in Sec.~\ref{losses}).

\subsection{VAE-conditioned texture generator}
\label{sec:VAE-conditioned}

\begin{figure}[t]
\centering
\includegraphics[width=0.7\linewidth]{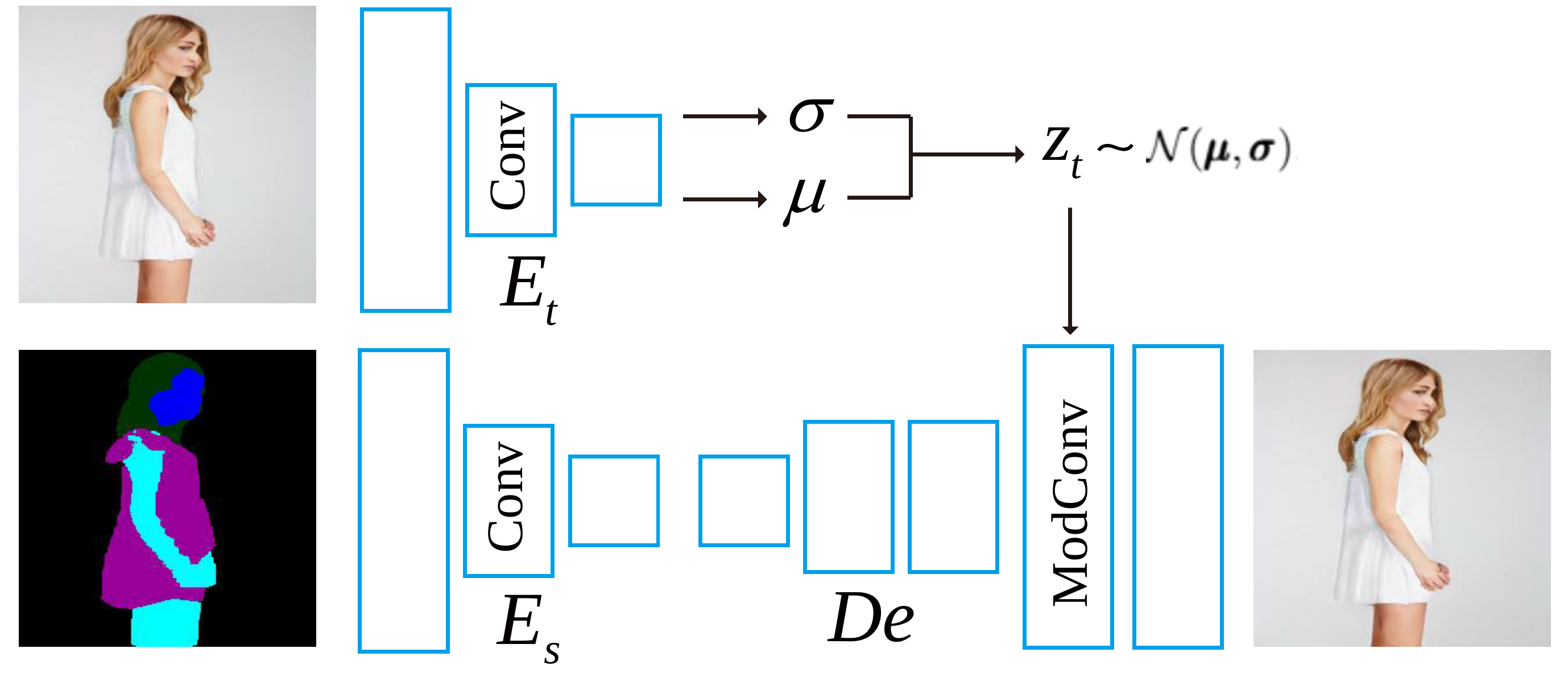}
\caption{The proposed VAE-conditioned texture generator. ModConv stands for ``Modulated Convolution" \cite{karras2020analyzing}.}
\label{fig:model2}
\end{figure}

The goal of our texture generator $G_{t}$ is twofold: (1) mapping the segmentation masks $\tilde I_s$ generated by $G_{3D}$ into a textured human image and (2) learning a marginal  distribution of the human texture using the dataset $T$.
The latter is obtained using a Variational AutoEncoder (VAE~\cite{kingma2013auto}) framework, which we use to learn how to {\em modulate} the texture style of the decoder.
Specifically, as shown in
Fig.~\ref{fig:model2},  $G_{t}$ is composed of a semantic encoder $E_{s}$, a texture encoder $E_{t}$, and a decoder $De$. 
$De$ is based on a
 StyleGANv2 architecture~\cite{karras2020analyzing}, in which a style code is used to ``demodulate" 
 the weights of each convolutional layer.
 We modify this architecture using a variational approach, in which the style code, {\em at inference time}, is extracted from a learned marginal distribution. In more detail,
 given a segmentation tensor $I_{s}$, we use $E_{s}$ to extract 
 the semantic content which  is decoded into the final image using $De$ and a texture code 
 $\pmb{z}_{t}$. The latter is sampled
using the VAE encoder $E_{t}$, which converts a real input image $I$ into a
latent-space normal distribution ($\mathcal{N}(\pmb{\mu},\pmb{\sigma})$), from which  $\pmb{z}_{t}$  is randomly chosen:
\begin{equation}
\label{eq.VAE}
\begin{aligned}
(\pmb{\mu},\pmb{\sigma}) = E_{t}(I), 
\pmb{z}_{t} \sim \mathcal{N}(\pmb{\mu},\pmb{\sigma}),\tilde I = De(E_s(I_s), \pmb{z}_{t}).
\end{aligned}
\end{equation}

$G_{t}$ is trained using the pairs in $T$. Specifically, given a pair of samples $(I^{i},I^{i}_{s})$, 
we use an adversarial loss $\ell^{t}_{adv}$ (and a dedicated discriminator $D_t$), jointly with a reconstruction loss $\ell_{r}$, and a standard Kullback-Leibler divergence ($\mathcal{D}_{kl}$) loss ($\ell_{kl}$) \cite{kingma2013auto}: 
\begin{equation}
\label{eq.rec-loss}
\ell_{r} = ||G_{t}(I^{i}_{s}, \pmb{z}_{t}) - I^{i}||_1, \ell_{kl} = \mathcal{D}_{kl}(\mathcal{N}(\pmb{\mu},\pmb{\sigma}) || \mathcal{N}(\pmb{0},\pmb{I})).
\end{equation}
Note that, in the reconstruction process, $G_{t}$ cannot ignore the segmentation tensor ($I^{i}_{s}$) and its corresponding encoder $E_{s}$. In fact, the information extracted from the real image $I$, and encoded using $E_{t}$, is not enough for the decoder to represent the  image content, since $\pmb{z}_{t}$ is used only as a style modulator in $De$.

\subsection{Consistency losses for semantics and pose disentanglement}
\label{consistency-losses}

In $G_{3D}$, the opacity value ($o$), computed by $h$, does not depend on the latent code $\pmb{z}_{s}$. Despite that, we have empirically observed that the semantics ($\pmb{z}_{s}$) and the pose ($\pmb{z}_{p}$) representations are highly entangled. 
We presume this is due to the convolutional filters in $R$ (\cref{Preliminaries}),
where the two latent factors are implicitly merged.
In order to increase the disentanglement of these factors,
 we propose two self-supervised consistency losses.

\begin{figure*}[!t] 
\centering
\includegraphics[width=1.0\linewidth]{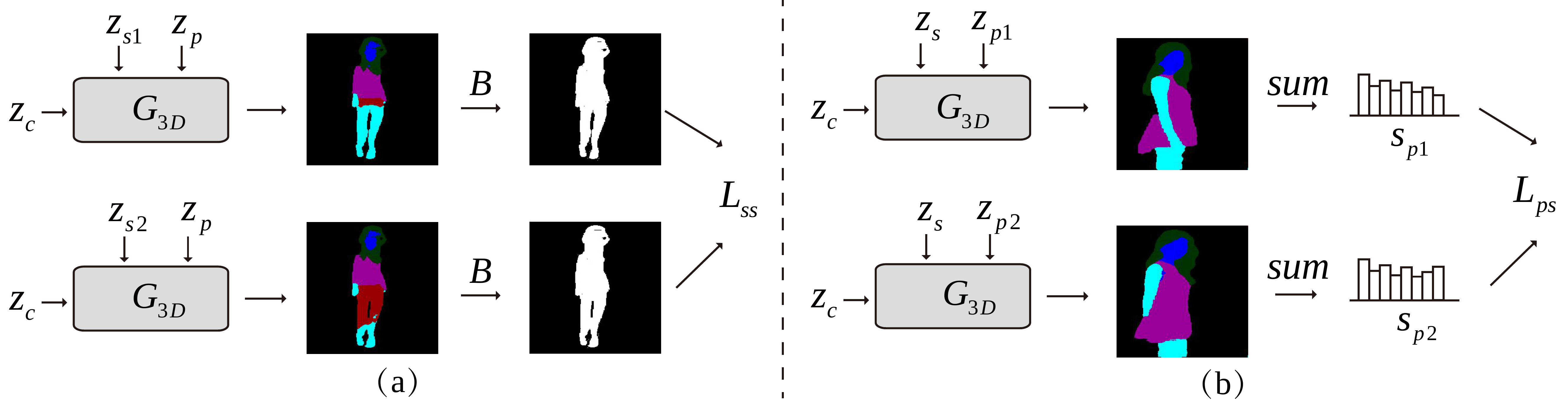}
\caption{A schematic representation of  our  consistency losses $\ell_{ss}$ (a) and $\ell_{ps}$ (b).}
\label{fig:model3}
\end{figure*}

\noindent \textbf{Silhouette-based geometric consistency.} This loss is based on the idea that two different
body segmentations
(e.g., long-sleeve vs. short- sleeve, etc.), produced using two different semantic codes $\pmb{z}_{s1}$ and $\pmb{z}_{s2}$, {\em but keeping fixed the pose and the camera codes},
  once they are binarized,
should correspond to roughly the same silhouette
 (see Fig.~\ref{fig:model3} (a)).
Formally, the proposed geometric consistency loss $\ell_{ss}$ is defined as:
\begin{equation}
\begin{aligned}
\ell_{ss} =& \Vert B(G_{3D}(\pmb{z}_{c}, \pmb{z}_{s1}, \pmb{z}_{p}) ) - B(G_{3D}(\pmb{z}_{c}, \pmb{z}_{s2}, \pmb{z}_{p})) \Vert_{1},
\end{aligned}
\end{equation}
where $B(\tilde I_s)$ maps  the segmentation masks $\tilde I_s$ into a binary silhouette image.

\noindent \textbf{Pose-based semantic consistency.}
Analogously to  $\ell_{ss}$, the proposed pose-based semantic consistency loss is based on the idea that 
two different  pose codes should produce a similar
 body segmentation. However, as shown in Fig.~\ref{fig:model3} (b), despite the body being partitioned in similar semantic segments (e.g., because the clothes have not changed), when the human pose changes, the overall spatial layout of these segments can also change (e.g., see the two different arm positions in  Fig.~\ref{fig:model3} (b)). For this reason, we formulate a semantic consistency loss ($\ell_{ps}$) which is spatial-invariant, and it is based on the channel-by-channel comparison of two segmentation masks. In more detail, 
 given two different pose codes $\pmb{z}_{p1}$ and $\pmb{z}_{p2}$, and fixing the semantics  and the camera code, we first produce two corresponding segmentation tensors $\tilde I_s^1$ and $\tilde I_s^2$. Then, 
 for each tensor and each channel, we
 sum all the channel-specific mask values over the spatial dimension   
 and we get 
 two spatial invariant vectors $\pmb{s}_{p1}, \pmb{s}_{p2} \in \mathbb{R}^{n_s}$. 
 Finally, $\ell_{ps}$ is given by:
\begin{equation}
\begin{aligned}
\ell_{ps} = \sum_{i=1}^{n_s}\lbrack {\rm max}(\frac{\mid \pmb{s}_{p1}[i] - \pmb{s}_{p2}[i] \mid}{\pmb{s}_{p1}[i] + \epsilon}, \rho) - \rho \rbrack,
\end{aligned}
\end{equation}
where $\pmb{s}[i]$ is the i-th channel value of vector $\pmb{s}$, 
$\epsilon$ is a small value used for numerical stability, and
 $\rho$ is a margin representing the tolerable channel-wise difference.

\subsection{Training and inference} 
\label{losses}

 $G_{3D}$ is trained using an adversarial loss
  ($\ell_{adv}^{s}$) jointly with $\ell_{ss}$ and $\ell_{ps}$ (Sec.~\ref{consistency-losses}):
\begin{equation}
\begin{aligned}
\ell_{3D} = \ell_{adv}^{s} + \lambda_{1} \ell_{ss} + \lambda_{2} \ell_{ps},
\end{aligned}
\end{equation}
where $\lambda_{1}$ and $\lambda_{2}$ are hyper-parameters  controlling  the contribution of each loss term.

$G_t$
is trained using a variational-adversarial  approach (VAE-GAN~\cite{10.5555/3045390.3045555}):
\begin{equation}
\label{eq.vae-loss}
\begin{aligned}
\ell_{tr} = \ell^{t}_{adv} + \lambda_{3} \ell_{r} + \lambda_{4} \ell_{kl},
\end{aligned}
\end{equation}
where $\lambda_{3}$ and $\lambda_{4}$ are hyper-parameters, and $\ell_{tr}$ is the overall objective function of $G_t$.

$G_{3D}$ and $G_t$ are trained separately. However, at inference time, the tensor $\tilde I_s$, generated by $G_{3D}$, is fed to $G_t$, along with a texture code $\pmb{z}_{t}$, randomly drawn from a standard normal distribution:
\begin{equation}
\begin{aligned}
\pmb{z}_{t} \sim \mathcal{N}(\pmb{0},\pmb{I}),\tilde I = De(E_{s}(\tilde I_s), \pmb{z}_{t}).
\end{aligned}
\end{equation}

\begin{figure*}[t] 
\centering
\includegraphics[width=1.0\linewidth]{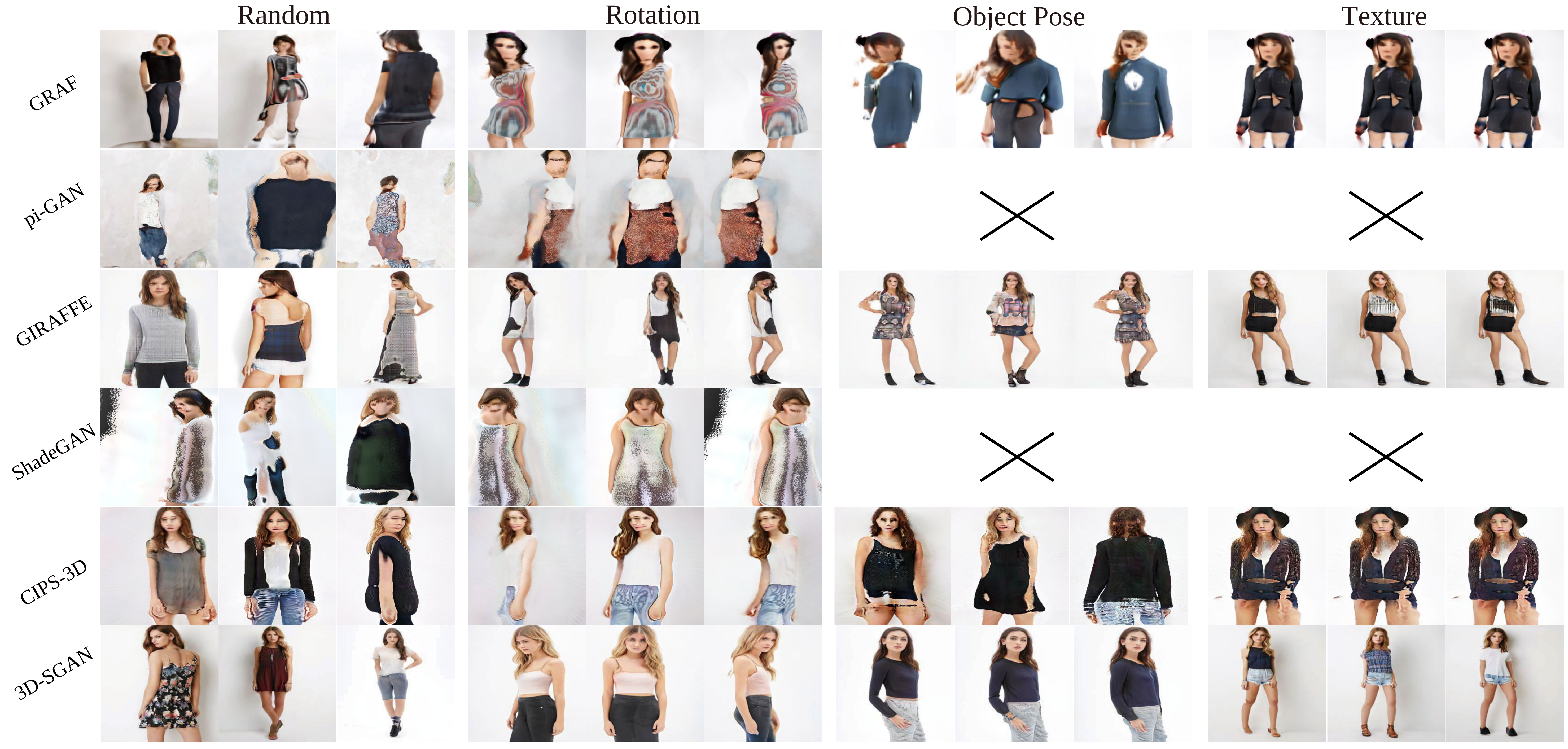}
\caption{A qualitative comparison. `Random' means that the results are generated by  randomly sampling the latent codes from the corresponding learned marginal distributions. The other 3 columns show controllable person generations  with respect to the rotation, the human pose, and the texture attribute. The lack of the
'Object Pose' and the 'Texture' results for 
both pi-GAN and ShadeGAN is due to the fact that both methods use a single latent code to model both the texture and the geometry.}
\label{fig:exp1}
\end{figure*}

\subsection{Real image editing using GAN inversion} 
\label{sec:inversion}
The variational method proposed in \cref{sec:VAE-conditioned} cannot completely reconstruct the input image. For real image editing, we use a GAN inversion technique~\cite{Abdal2019Image2StyleGANHT} 
to optimize the values of the latent codes corresponding to a real input image $I$. Since we have two separate generators ($G_{3D}$ and $G_t$), the optimization process is based on two steps (see Fig.~\ref{fig:model}, bottom).
Specifically, 
given a pair of real image and its corresponding segmentation masks (extracted using \cite{Badrinarayanan2017SegNetAD}, see Sec.~\ref{G3D}) $(I, I_s)$, we first generate  $\tilde I_s = G_{3D}(\pmb{z}_c, \pmb{z}_s, \pmb{z}_p)$ and we optimize 
$|| \tilde I_s - I_{s}||_1$ with respect to $\pmb{z}_c, \pmb{z}_s$ and $\pmb{z}_p$.
Let $\pmb{z}_c^*, \pmb{z}_s^*$ and $\pmb{z}_p^*$ be the optimal values so found, and let $\tilde I_s^* = G_{3D}(\pmb{z}_c^*, \pmb{z}_s^*, \pmb{z}_p^*)$.
 Then, we use $\tilde I =G_t(\tilde I_s^*, \pmb{z}_t)$ 
 and we optimize $|| \tilde I - I||_1 + \tau LPIPS(\tilde I, I)$ with respect to  $\pmb{z}_t$,
 where $LPIPS(I^1, I^2)$ is the $LPIPS$ distance between two images~\cite{zhang2018perceptual} 
 and we use $\tau = 10$.

Once obtained the latent codes ($\pmb{z}_c^*, \pmb{z}_s^*, \pmb{z}_p^*, \pmb{z}_t^*$) corresponding to a real image, editing can be easily done by changing these codes.

\section{Experiments} 
\label{exper}

\begin{table*} [t]
\caption{A quantitative comparison using the FID  ($\downarrow$) and the aMP ($\uparrow$) scores. 
}
\label{tab:quan1}
\centering
\resizebox{0.8\linewidth}{!}{
\begin{tabular}{lclclclclcl}
\toprule
\multirow{2}{*}{Method} & \multicolumn{4}{c}{FID $\downarrow$} & aMP $\uparrow$ \\
\cmidrule(r){2-5}
\cmidrule(r){6-6}
 & Random & Rotation & Object-Pose & Texture & Rotation \\
\midrule
GRAF~\cite{Schwarz2020NEURIPS} & 52.68 & 176.9 & 57.76 & 220.9 & 64.0 \\
pi-GAN~\cite{chanmonteiro2020pi-GAN} & 137.6 & 213.7 & - & - & 58.0  \\
GIRAFFE~\cite{GIRAFFE} & 42.73 & 123.4 & 82.61 & 98.41 & 51.0 \\
ShadeGAN~\cite{pan2021shading} & 134.7 & 232.4 & - & - & {\bfseries 89.0} \\
CIPS-3D~\cite{zhou2021CIPS3D} & 69.45 & 156.9 & 233.6 &  {\bfseries36.16} & 60.0  \\
3D-SGAN & {\bfseries 8.240} & {\bfseries 117.3} & {\bfseries 54.00} &  60.63 & 81.0 \\
\bottomrule
\end{tabular}}
\end{table*}

\begin{figure*}[t] 
\centering
\includegraphics[width=1.0\linewidth]{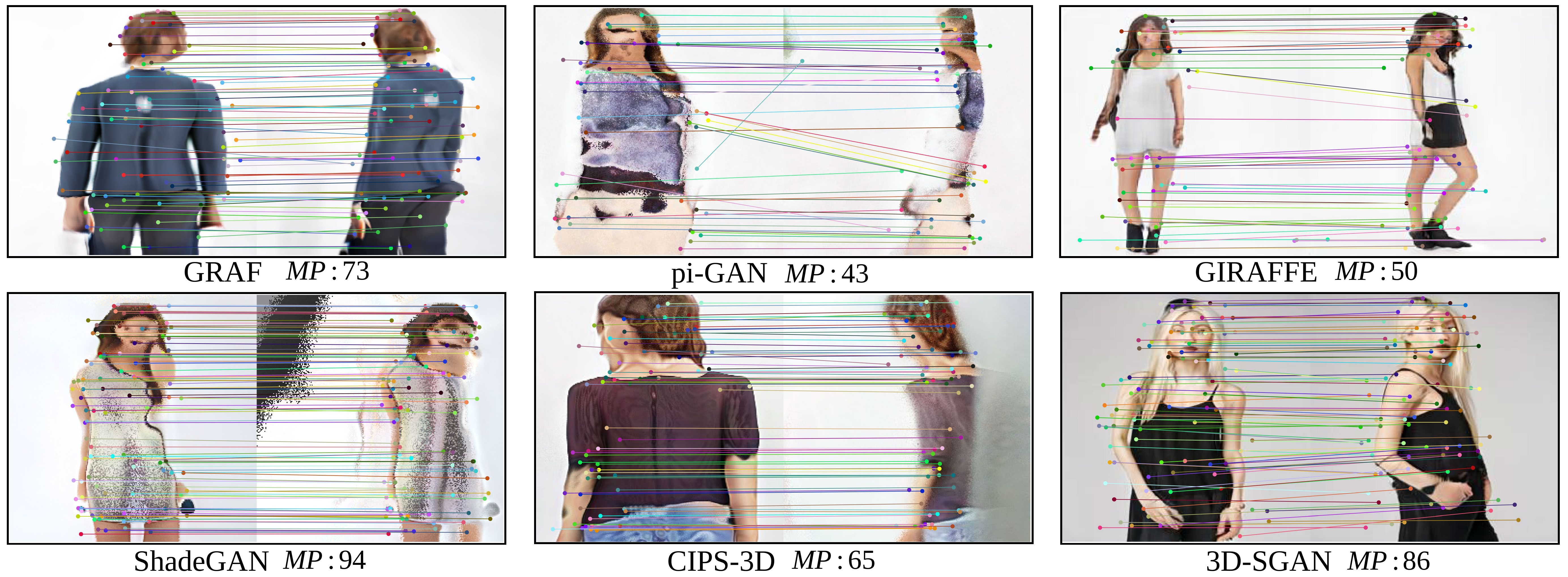}
\caption{Computing $MP$ between pairs of generated images with 2 different viewpoints.}
\label{fig:exp2}
\end{figure*}

\noindent \textbf{Datasets.} 
We use the DeepFashion In-shop Clothes Retrieval benchmark~\cite{liu2016deepfashion}, which consists of 52,712 high-resolution ($1101{\times}750$ pixels) person images  with various appearances and poses. This dataset has been widely used in  pose transfer tasks. We use the following preprocessing. First, we remove overly cropped images, such as incomplete images of humans. Then, the remaining 42,977 images are resized into a  $256{\times}256$ resolution, and are divided into 41,001 training  and 1,976 testing images. 

\noindent \textbf{Training details.} 
Following GIRAFFE~\cite{GIRAFFE}, the camera distribution $P_{c}$ can be implemented by first sampling the camera code from a uniform distribution over the dataset-dependent camera elevation angles, and then applying an object affine transformation to sample 3D points and rays. Both $G_t$ and  $G_{3D}$ are trained using the RMSprop optimizer~\cite{Kingma2014AutoEncodingVB}. The learning rate for both the discriminator and the generator is set to $10^{-4}$. For the loss weights, we use: $\lambda_{1}{=}0.01$, $\lambda_{2}{=}0.01$, $\lambda_{3}{=}1$, and $\lambda_{4}{=}1$. For GAN inversion, we use the Adam optimizer \cite{Kingma2015AdamAM} with a learning rate of $10^{-2}$.

\noindent \textbf{Baselines.}
We compare 3D-SGAN with  five state-of-the-art 3D-aware generative approaches, i.e., GRAF~\cite{Schwarz2020NEURIPS}, pi-GAN~\cite{chanmonteiro2020pi-GAN}, GIRAFFE~\cite{GIRAFFE}, ShadeGAN~\cite{pan2021shading} and CIPS-3D~\cite{zhou2021CIPS3D}. For each baseline, we use the corresponding publicly available code with a few minor adaptations for the DeepFashion dataset. Note that some concurrent methods, such as  StyleNeRF~\cite{gu2021stylenerf}, GRAM~\cite{deng2021gram}, Tri-plane~\cite{chan2021efficient}, StyleSDF~\cite{or2021stylesdf} achieve a performance very similar to CIPS-3D. Moreover, for some of them there is no released  code yet,  thus, a direct comparison is not possible. The comparison with the 2D-GAN model HumanGAN~\cite{sarkar2021humangan} can be found in the supplementary material~\ref{sec:sm_result}.

\noindent \textbf{Metrics.}
We adopt the widely used FID~\cite{NIPS2017_7240} scores to evaluate the quality of the generated human images, following common protocols (e.g., using  5,000  fake samples, etc.). And we propose the average Matched Points (aMP) to evaluate the 3D-view consistency of the generated images. The supplementary document provides the introduction of this metric.

\begin{figure*}[t]
\centering
\includegraphics[width=1.0\linewidth]{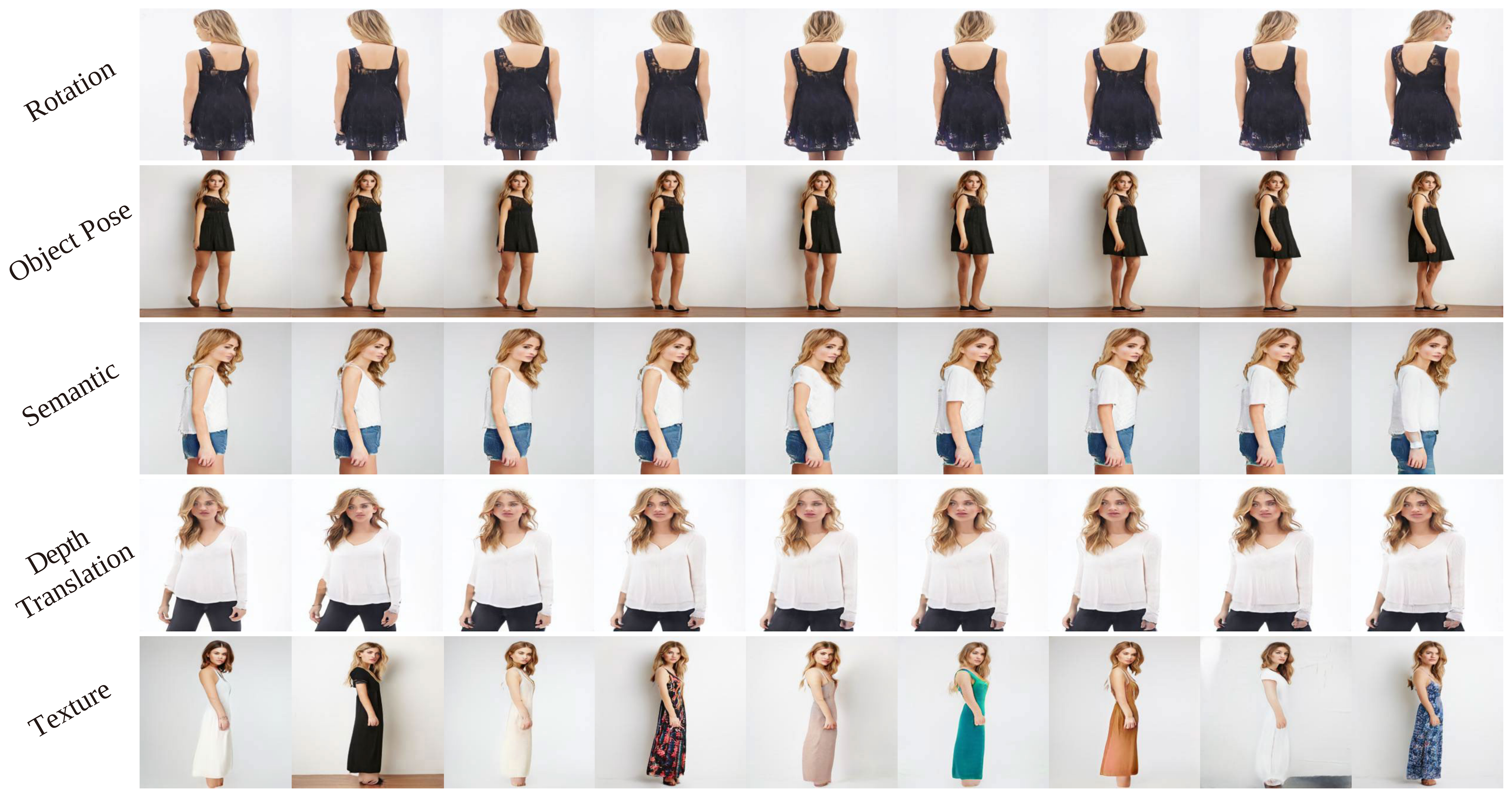}
\caption{Controllable person generation by interpolating latent codes (Rows 1-4). The fifth row shows texture generation results obtained  randomly sampling $\pmb{z}_{t}$.}
\label{fig:exp12}
\end{figure*}

\subsection{Comparisons with state-of-the-art methods} \label{quaquan}

\noindent \textbf{Unconditioned human generation.}
Fig.~\ref{fig:exp1} (``Random'' column) shows a qualitative comparison between  image samples generated by all the models. GRAF~\cite{Schwarz2020NEURIPS}, pi-GAN~\cite{chanmonteiro2020pi-GAN} and ShadeGAN~\cite{pan2021shading} fail to generate realistic human images. GIRAFFE~\cite{GIRAFFE} and CIPS-3D~\cite{zhou2021CIPS3D} generate  reasonable human images, but they  suffer from visual artifacts and texture blurs. In contrast, 3D-SGAN synthesizes much better and more photo-realistic images. 
This qualitative analysis is confirmed in 
 Tab.~\ref{tab:quan1}, where the corresponding FID scores show that   3D-SGAN significantly outperforms all the other baselines. 

\noindent \textbf{Controllable human generation.}
We analyse the representation controllability of all the models, which reflects the ability  to disentangle different attributes from each other. We do this by manipulating a single latent code while fixing the others.
Fig.~\ref{fig:exp1} (columns ``Rotation", ``Object Pose" and ``Texture") shows a qualitative comparison by varying  only a single latent code. We  observe that all the models can rotate the camera viewpoint. However, GRAF and CIPS-3D fail to disentangle the object pose and the texture. Moreover, pi-GAN and ShadeGAN also suffer from the same problem, since they use one single latent code to model both texture and geometry. 
On the other hand, both GIRAFFE~\cite{GIRAFFE} and 3D-SGAN can effectively disentangle the different variation factors, but GIRAFFE~\cite{GIRAFFE} suffers from multi-view inconsistencies and mode collapse for the texture generation. 
In Table~\ref{tab:quan1}, we use FID scores to evaluate the realistic degree of each attribute (e.g., “Rotation”, etc.). This is done computing FIDs using only the manipulated (e.g., rotated) fake images, which are compared with all the real images in the dataset. Note that this protocol cannot measure the attribute-based consistency.
In most cases, 3D-SGAN has better FID scores than the other baselines. 

In order to evaluate the 3D-view consistency, we use our proposed aMP metric (\cref{exper}).
Table~\ref{tab:quan1} shows that 3D-SGAN gets the best aMP scores with respect to 
all the  other methods except from ShadeGAN, which however generates much less realistic images, as testified by the very high FIDs (134.7 vs. our 8.24, Table~\ref{tab:quan1}, first column) and qualitatively shown in  Fig.~\ref{fig:exp2}.

Fig.~\ref{fig:exp12} shows additional controllable human image generation results obtained with  3D-SGAN. The generated images are realistic and, in most cases, the attributes are effectively disentangled. Specifically, Fig.~\ref{fig:exp12} (1-st row) shows camera rotation results. The images generated by interpolating the camera pose parameter are consistent, and the transition from one image to the next is smooth, while simultaneously preserving the other attributes such as the  texture and the pose. On the other hand, the second row shows images generated by interpolating the pose code. We again observe that human identity has been well preserved. Similarly, the other rows  show that the non-target attributes have been well preserved. Finally, the third row shows that the head poses from left to right undergo only minor changes (``face frontalization''). This is likely due to both the limited training data and  the  data bias of the typical  fashion images, where people have a frontal face. Additional results are shown in the Supplementary Material~\ref{sec:sm_result}. 

\begin{figure*}[t] 
\centering
\includegraphics[width=1.0\linewidth]{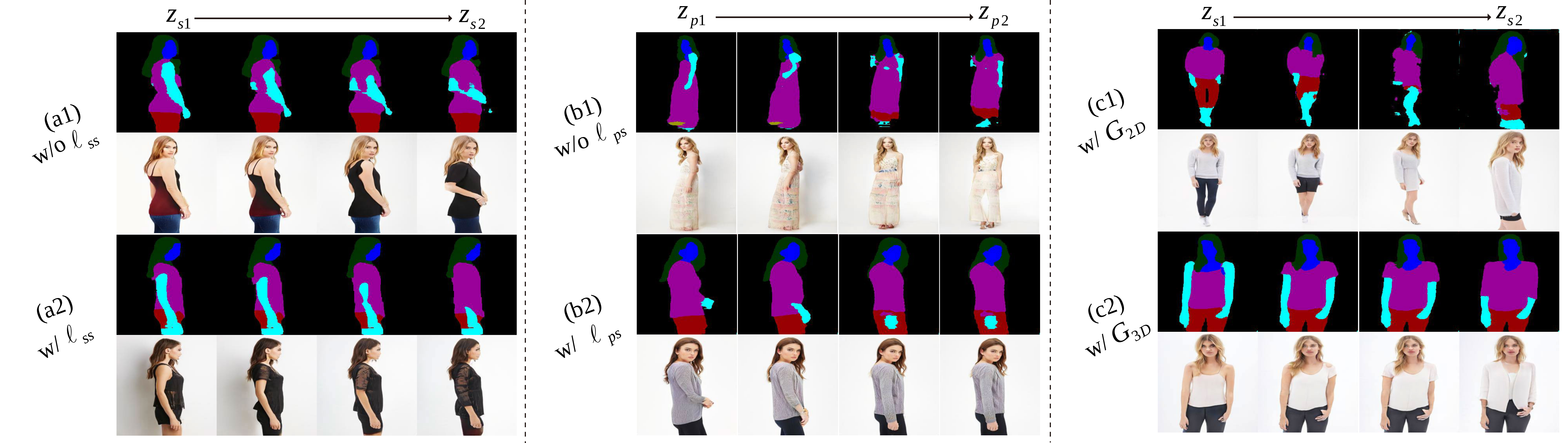}
\caption{A qualitative analysis of  $\ell_{ss}$ (a),  $\ell_{ps}$ (b), and $G_{3D}$ (c). (a) and (c) show  interpolation results between semantics codes $\pmb{z}_{s1}$ and $\pmb{z}_{s2}$. (b) shows  interpolation results between pose codes $\pmb{z}_{p1}$ and $\pmb{z}_{p2}$.}
\label{fig:ab1}
\end{figure*}

\begin{table}[t] 
\caption{A quantitative analysis of  $\ell_{ss}$ (left) and $\ell_{ps}$ (right).
In the latter case,  we use LPIPS to measure the diversity of sample pairs generated by interpolating $\pmb{z}_p$.} 
\centering
\begin{tabular}{ccc|ccc}
\toprule
Metrics & w/o $\ell_{ss}$  & w/ $\ell_{ss}$ & Metrics & w/o $\ell_{ps}$  & w/ $\ell_{ps}$  \\
\midrule
L1 $\downarrow$  & 5.2489 & \textbf{3.9614} & LPIPS $\downarrow$  &  0.1132 & \textbf{0.0393} \\
\bottomrule
\end{tabular}
\label{tab:ab_quan1}
\end{table}

\begin{figure*}[t] 
\centering
\includegraphics[width=1.0\linewidth]{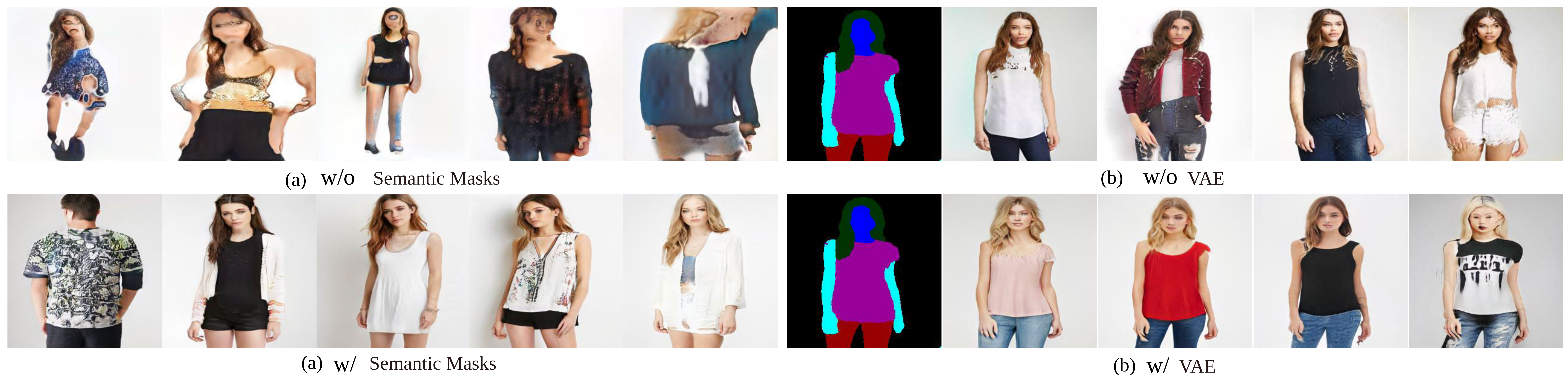}
\caption{An analysis of the impact of the  semantic masks (a) and the VAE-conditioned texture generator (b).}
\label{fig:ab3}
\end{figure*}

\subsection{Ablation study}

\begin{table}[t] 
\caption{A quantitative analysis of the 3D generator $G_{3D}$, the semantic masks (SMs) and the VAE in our 3D-SGAN.} 
\centering
\begin{tabular}{ccccc}
\toprule
Metrics & w/ $G_{2D}$  & w/o SMs & w/o VAE  & full \\
\midrule
FID $\downarrow$  & 13.24 & 66.79 &  14.35 & \textbf{8.240} \\
\bottomrule
\end{tabular}
\label{tab:ab_quan2}
\end{table}

\noindent \textbf{The consistency losses.}
 Fig.~\ref{fig:ab1} (a) shows a comparison between the results generated by 3D-SGAN
with  and without $\ell_{ss}$. The effectiveness of $\ell_{ss}$ is shown by observing that, when removed, the generation process suffers from serious geometric inconsistencies. Specifically, the segmentation masks in Fig.~\ref{fig:ab1} (a1) have undesirable pose variations, while Fig.~\ref{fig:ab1} (a2) shows that $\ell_{ss}$ can largely alleviate this problem. To quantitatively evaluate this effect, we randomly sample two different semantic codes and we compute the $L1$ distance between the silhouettes of the corresponding generated segmentations. We average the scores over 500 different samples. The results  reported in Table~\ref{tab:ab_quan1} (left) validate the effectiveness of this loss for improving the geometric consistency. 

Analogously, Fig.~\ref{fig:ab1} (b) qualitatively evaluates the impact of $\ell_{ps}$  with respect to the semantic consistency over different pose codes. 
For instance, in Fig.~\ref{fig:ab1} (b1)  there is no ``red'' region in the segmentation masks in the first and in the second column. However, this region is present in  columns 3 and 4. Conversely, Fig.~\ref{fig:ab1} (a2) shows  that  $\ell_{ps}$ can alleviate this phenomenon. To quantitatively evaluate $l_{ps}$, we use LPIPS~\cite{zhang2018perceptual},  and we measure the average
{\em pairwise}
diversity of the sample pairs generated by interpolating $z_p$ (the lower the diversity, the higher the intra-pair consistency). Tab.~\ref{tab:ab_quan1} (right) shows that the full model achieves a lower diversity than the variant without $\ell_{ps}$.

\noindent \textbf{The 3D generator.}
To evaluate the benefit of using a GNeRF based generator, we replace it with a vanilla GAN ($G_{2D}$), which takes the  pose  and the semantics code as inputs. In this experiment, we keep all the other modules fixed. Note that $G_{2D}$ cannot manipulate the camera parameters and, thus, it cannot generate images from multiple viewpoints. Moreover, $G_{3D}$ can  better disentangle the semantics and the pose factors with respect to $G_{2D}$, as demonstrated by Fig.~\ref{fig:ab1} (c), where we show interpolation results between two different semantic codes. Tab.~\ref{tab:ab_quan2} shows the $G_{3D}$ (the full model) achieves significantly better FID scores than $G_{2D}$.

\noindent \textbf{The semantic masks and the texture generator.} Existing methods such as GRAF~\cite{Schwarz2020NEURIPS} and GIRAFFE~\cite{GIRAFFE} do not use an additional texture generator which translates semantic masks into textured images. In contrast, the effectiveness of our semantic-based approach is shown in Fig.~\ref{fig:exp1} and Table~\ref{tab:quan1}. However, to provide an apple-to-apple comparison and further verify the effectiveness of the semantic masks, we use an additional baseline. Specifically, in this baseline, we render the 3D representation of $G_{3D}$ into features rather than semantic masks and we use the texture generator to map these features into the final image. Fig.~\ref{fig:ab3} (a) shows the comparison of our full model with this baseline. We observe that the baseline (w/o semantic masks) fails to generate high-quality human images. 
Tab.~\ref{tab:ab_quan2} shows that the full model
 quantitatively outperforms this baseline  in terms of FID scores.

\noindent \textbf{The Variational Autoencoder.} We evaluate the effect of conditioning $G_t$ using a VAE (\cref{sec:VAE-conditioned}). This is done by removing the texture encoder $E_t$ 
jointly with $\ell_{kl}$ and  $\ell_{r}$ from \cref{eq.vae-loss}.
Fig.~\ref{fig:ab3} (b) shows the comparison between the VAE-based approach and this variant. Both models generate human images with a high texture variability. However, the variant w/o VAE fails to preserve semantic information, i.e., the coherence between the semantic masks, describing the clothes layout, and the final generated clothes. This shows that our VAE-based $G_t$ learns to effectively map the semantic tensors to human images while modeling the texture distribution with the latent code $\pmb{z}_{t}$. 
Tab.~\ref{tab:ab_quan2} shows that this variant
is significantly outperformed by the 
 proposed VAE-based encoder.
 
\subsection{Real human image editing}
\begin{figure}[!t] 
\centering
\includegraphics[width=1.0\linewidth]{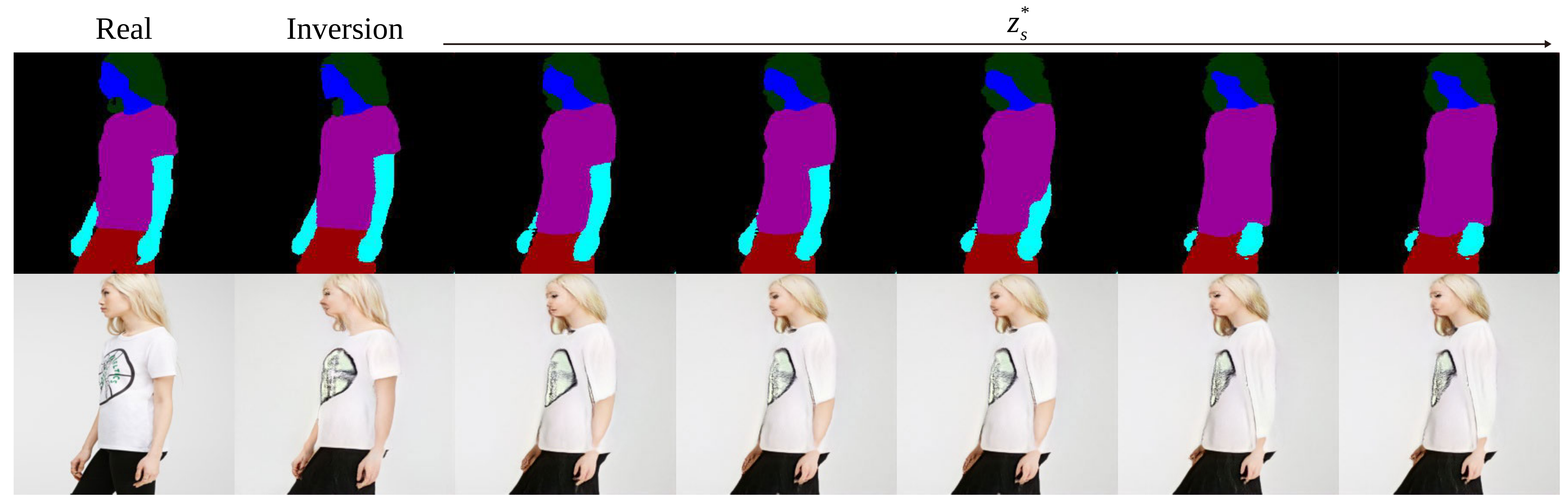}
\caption{Real data semantic editing results using GAN inversion.}
\label{fig:real}
\end{figure}

In this section, we use  GAN inversion   for real data editing tasks. The second column of Fig.~\ref{fig:real} shows that 
the optimal  code values $(\pmb{z}_c^*, \pmb{z}_s^*, \pmb{z}_p^*, \pmb{z}_t^*)$, obtained using the procedure described in Sec.~\ref{sec:inversion}, lead to an effective reconstruction of the real input data (first column).
In the other columns, 
 we linearly manipulate the semantic code $\pmb{z}_s^*$ while keeping fixed the other codes.
 Specifically, the second row of Fig.~\ref{fig:real} shows the generated images corresponding to the semantic masks in the first row. These results demonstrates the effectiveness of the GAN inversion mechanism and the possibility to apply our model to a wide range of human image editing tasks.
 We computed the average
LPIPS and MS-SSIM scores between real and inversion images,
respectively obtaining 0.0301 and 0.912, which confirms
the high reconstruction quality of our inversion.

\section{Conclusion}


We proposed a  3D-aware Semantic-Guided Generative model (3D-SGAN) for human synthesis. We use a generative NeRF to implicitly represent the 3D human body and we render the 3D representation into 2D segmentation masks. Then, these masks are mapped into the final  images using a VAE-conditioned texture generator. Moreover, we propose two consistency losses further disentangle the  pose and the semantics factors. Our experiments show that the proposed approach generates human images which are significantly more realistic and more controllable than state-of-the-art methods.

\noindent \textbf{Acknowledgements}
This work was supported by the EU H2020 projects AI4Media (No.951911) and SPRING (No. 871245). 

\clearpage
%
%
\bibliographystyle{splncs04}
\bibliography{3dsgan}

\section{Supplementary Material} \label{sec:sm}

This section supplements our paper 3D-Aware GAN by providing more details of the implementation, training, the proposed metric as well as the additional experimental results on DeepFashion and another VITON dataset. Finally, we provide the detailed discussion about the training strategy and limitations of the model.

\subsection{Implementation details}
\label{sec:sm_details}

\textbf{Training.} Besides the training details reported in Sec. 5  of the paper,  here we add that, in order to stabilize the training of $G_{3D}$,  the loss weight $\lambda_2$ controlling the influence of  $\ell_{ps}$ (Sec. 4.4 of the main paper) is set to 0 for the initial 100,000 training steps. 

\textbf{Dataset-specific parameters.} The field of view is 10$^{\circ}$ for both datasets (DeepFashion and VITON). The camera elevation is 10$^{\circ}$ on both datasets, and the object rotation is 360$^{\circ}$ for DeepFashion and 72$^{\circ}$ for VITON. We estimated these parameters from the empirical distribution of each dataset. More details about the parameters and the network architecture can be found in our code. For the baselines, \textit{i.e.,} GRAF~\cite{Schwarz2020NEURIPS}, pi-GAN~\cite{chanmonteiro2020pi-GAN}, GIRAFFE~\cite{GIRAFFE}, ShadeGAN~\cite{pan2021shading} and CIPS-3D~\cite{zhou2021CIPS3D}, we used their publicly available code and we trained all the models using  the architecture and the configuration corresponding to  CelebA dataset~\cite{Liu2015DeepLF}. The details of the CelebA configuration can be found in  corresponding paper or in the public code of each baseline. 
However, for a fair comparison, we changed a few baseline parameters, such as the field of view, to be consistent with  our model.

\textbf{Average Matched Points (aMP).} And we propose the average Matched Points (aMP) to evaluate the 3D-view consistency of the generated images based on local region matching. Specifically, we use Patch2Pix~\cite{zhou2021patch2pix} to compute a  point-wise matching between two generated images 
($\tilde I_1, \tilde I_2$) of the same person with different viewpoints, then we count the number of Matched Points $MP(\tilde I_1, \tilde I_2)$. $MP$ is applied to image pairs with the same identity (texture) but different rotation angles. In more detail,
for each method, 
we randomly generate 500 samples by varying the texture content  while keeping fixed the other variation factors. Then, for each of these 500 samples, we change the camera viewpoint in order to get 3 different random rotations. We can now compute $MP(\tilde I_1, \tilde I_2)$ for all the 6 possible pairwise combinations of these 3 samples (note that applying Patch2Pix to $(\tilde I_1, \tilde I_2)$ gets slightly different results from $(\tilde I_2, \tilde I_1)$). Finally, we average $MP$ over the $500 \times 6$ pairs and we get a score which we call 
{\em average MP score} (aMP).

\subsection{Additional results}
\label{sec:sm_result}

 \textbf{The VITON dataset.}
In addition to the DeepFashion dataset~\cite{liu2016deepfashion}, used in the main paper, we also compare our method with the baselines on the VITON dataset~\cite{han2017viton}. VITON is composed of 16,253 front-view woman and top-body clothes image pairs and it is widely used for virtual try-on tasks. We use the front-view woman images and we divide the dataset into 14,221 training images and 2,032 testing images. The original image resolution is $176 \times 256$, but we resize all the images to $256 \times 256$.

\begin{figure*}[t]\small
\begin{center}
\includegraphics[width=1.0\linewidth]{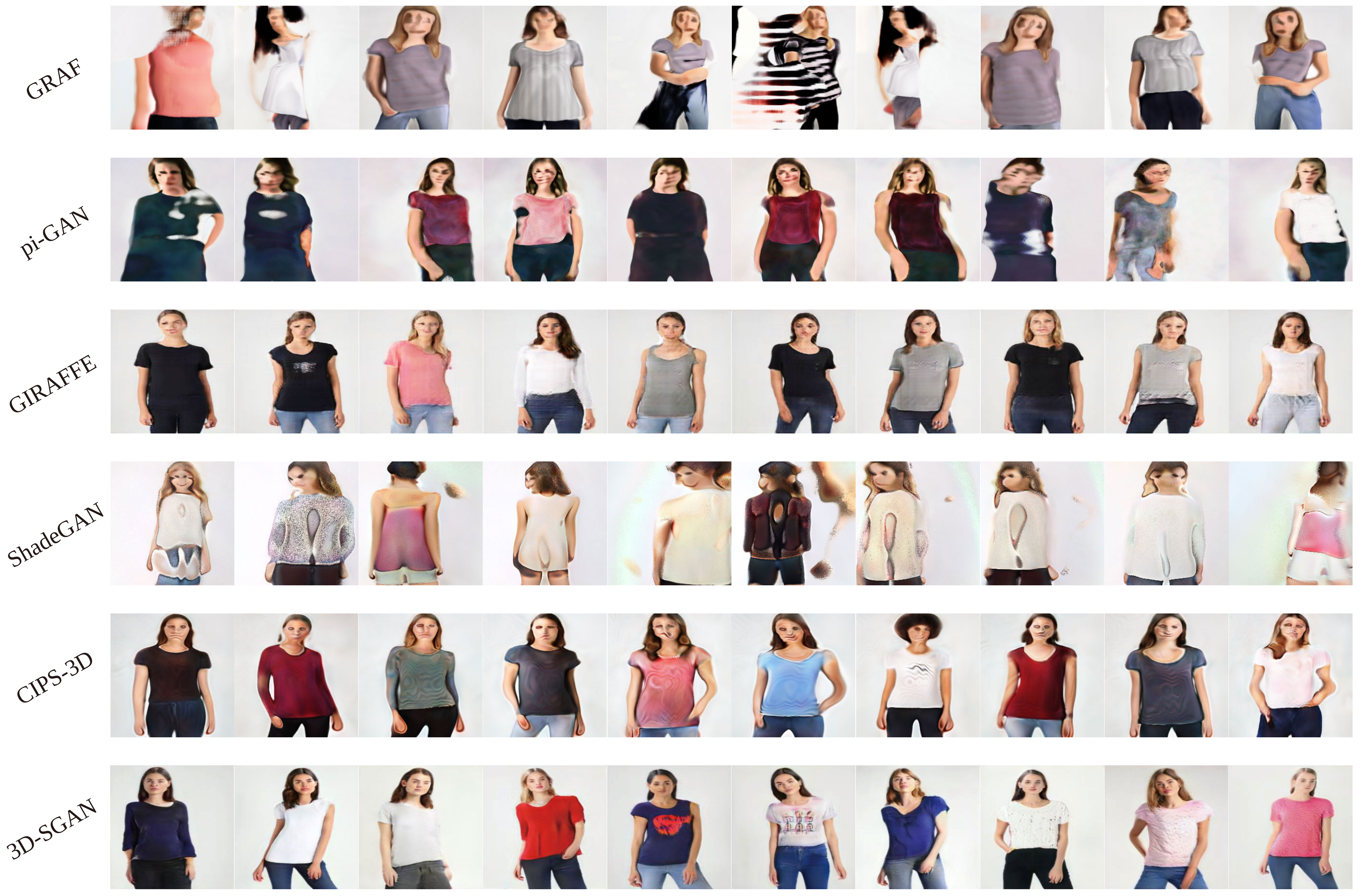}
\end{center}
\caption{Unconditioned human image generation. A comparison between our 3D-SGAN with all the baselines using the VITON dataset. }
\label{fig:sm1}
\end{figure*}

\begin{figure*}[t]\small
\begin{center}
\includegraphics[width=1.0\linewidth]{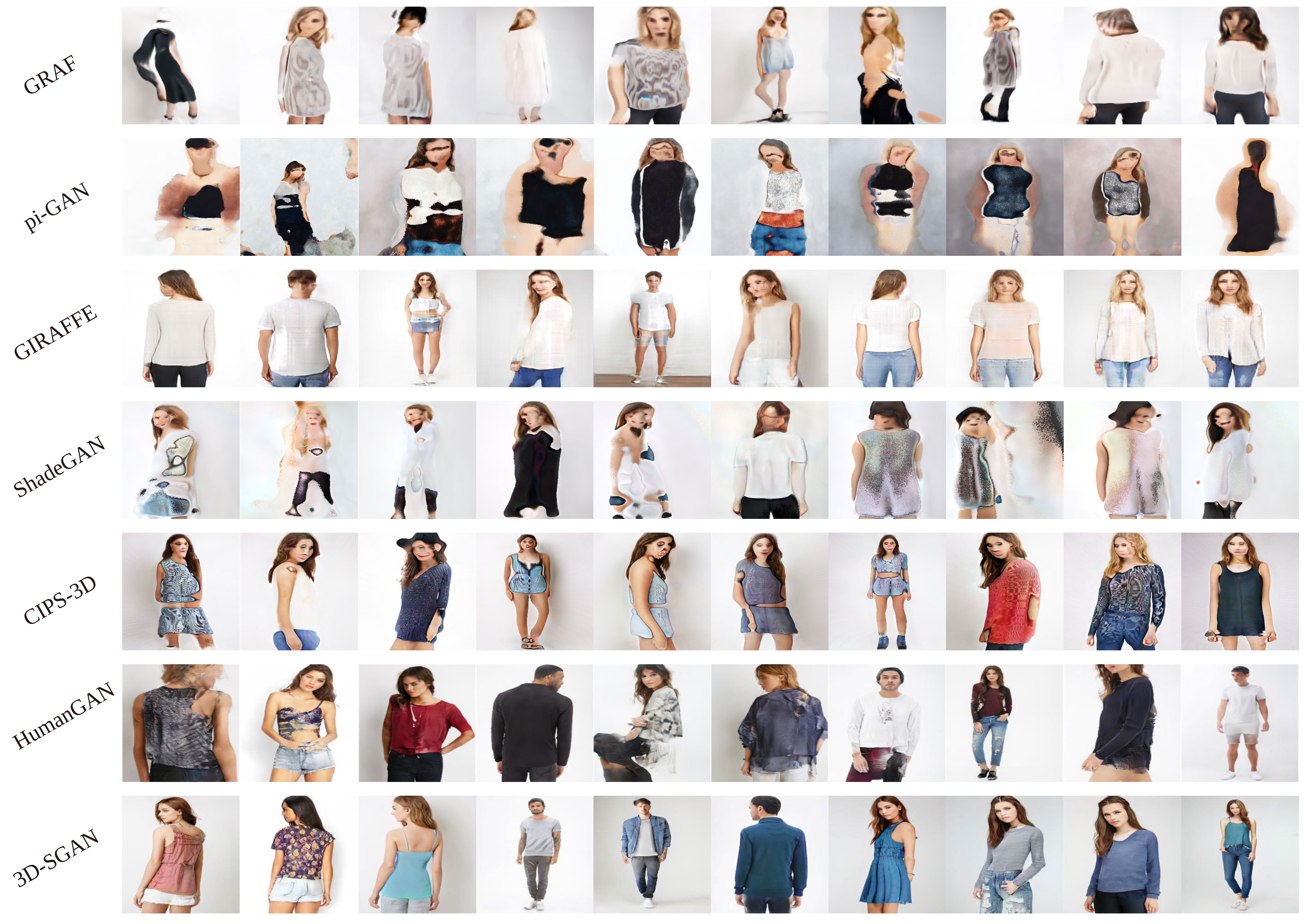}
\end{center}
\caption{Unconditioned human image generation. A comparison between our 3D-SGAN with all the baselines using the  DeepFashion dataset.}
\label{fig:sm2}
\end{figure*}

 \textbf{Unconditioned human image generation.}
Fig.~\ref{fig:sm1} and Fig.~\ref{fig:sm2} show a qualitative comparison between image samples generated by all the models using both the datasets. Our method generates more realistic human images than all the other baselines on both datasets. Note that our model achieves  results comparable with HumanGAN~\cite{sarkar2021humangan} in human generation on the DeepFashion dataset. However, HumanGAN is not a 3D-Aware model, and it fails to control 3D factors. Note that HumanGAN~\cite{sarkar2021humangan} takes a source sample as input, from which most of the target appearance can be copied in the training stage.
Conversely, our images are generated from noise, which is an harder task.
Moreover, despite HumanGAN is trained on DeepFashion, there is no public code for training, thus for our comparison we had to use the sample images available in the  official Web page (which are not generated unconditionally from noise, so probably the above comparison is a bit in favor of HumanGAN).

\begin{figure*}[t]\small
\begin{center}
\includegraphics[width=1.0\linewidth]{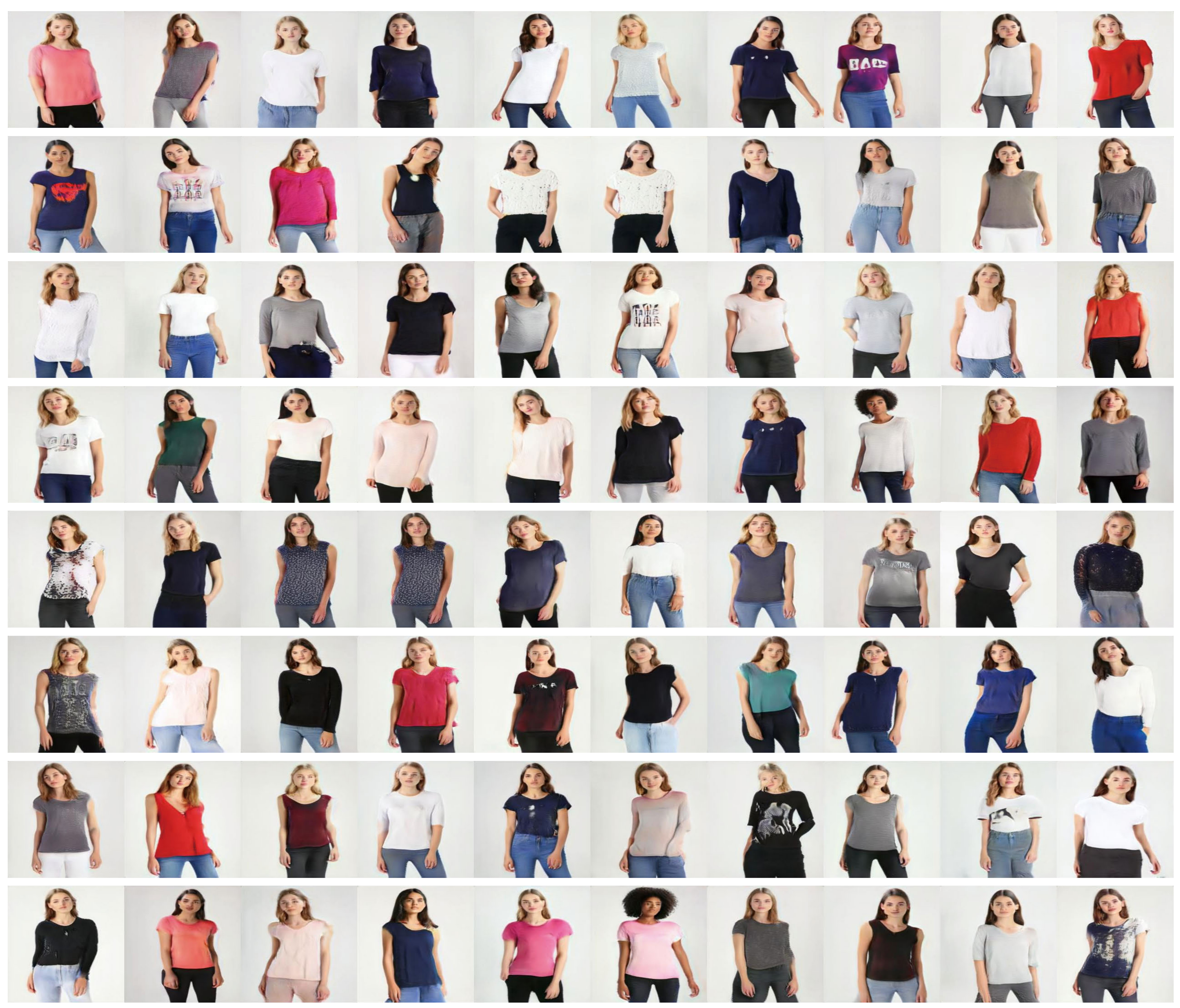}
\end{center}
\caption{VITON dataset: additional randomly generated images using our method.}
\label{fig:sm9}
\end{figure*}

\begin{figure*}[t]\small
\begin{center}
\includegraphics[width=1.0\linewidth]{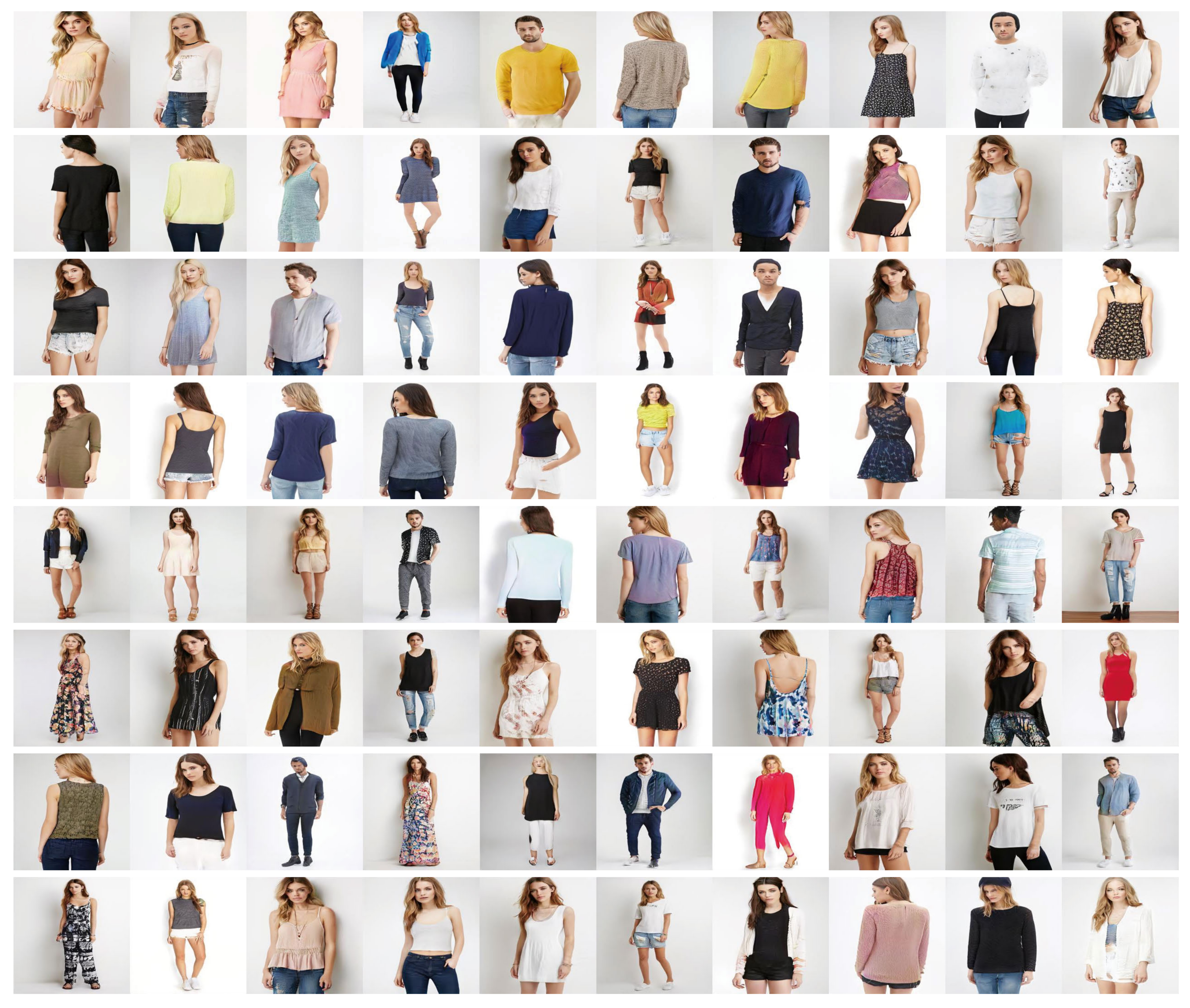}
\end{center}
\caption{Deepfashion dataset: additional randomly generated images using our method.}
\label{fig:sm10}
\end{figure*}

The quantitative evaluation provided in Tab.~\ref{tab:quan}
Tab.~\ref{tab:quan1}  using the user studies and FID scores, shows that our approach significantly outperforms all the other methods also on the DeepFashion and VITON dataset.  Fig.~\ref{fig:sm9} and Fig.~\ref{fig:sm10} show additional image generation results on both datasets, obtained using our method.

\begin{table}[!h] 
\centering
\caption{User studies on the DeepFashion dataset.}
\resizebox{1.0\linewidth}{!}{%
\begin{tabular}{lcccccccc} 
\toprule
Method &  3D-SGAN & HumanGAN & GIRAFFE & CIPS-3D & pi-GAN & ShadeGAN & GRAF \\ \midrule
User Study $\uparrow$  &  {\bf 55.2} \% & 41.2\% & 3.2\% & 0.2\% & 0.05\% & 0.0\% & 0.0\% \\
\bottomrule
\end{tabular}}

\label{tab:quan}
\end{table}

\begin{table}[t] 
\centering
\caption{A quantitative comparison using FID scores on the VITON dataset.}
\resizebox{1.0\linewidth}{!}{%
\begin{tabular}{ccccccc}
\toprule
Method & GRAF~\cite{Schwarz2020NEURIPS} & pi-GAN~\cite{chanmonteiro2020pi-GAN} & GIRAFFE~\cite{GIRAFFE} & ShadeGAN~\cite{pan2021shading} & CIPS-3D~\cite{zhou2021CIPS3D} & 3D-SGAN (Ours) \\
\midrule
FID $\downarrow$  & 67.300 & 121.15 & 26.750 & 110.02 & 48.919 & {\bfseries 14.060} \\
\bottomrule
\end{tabular}}
\label{tab:quan1}
\end{table}

\begin{figure*}[t]
\begin{center}
\includegraphics[width=1.0\linewidth]{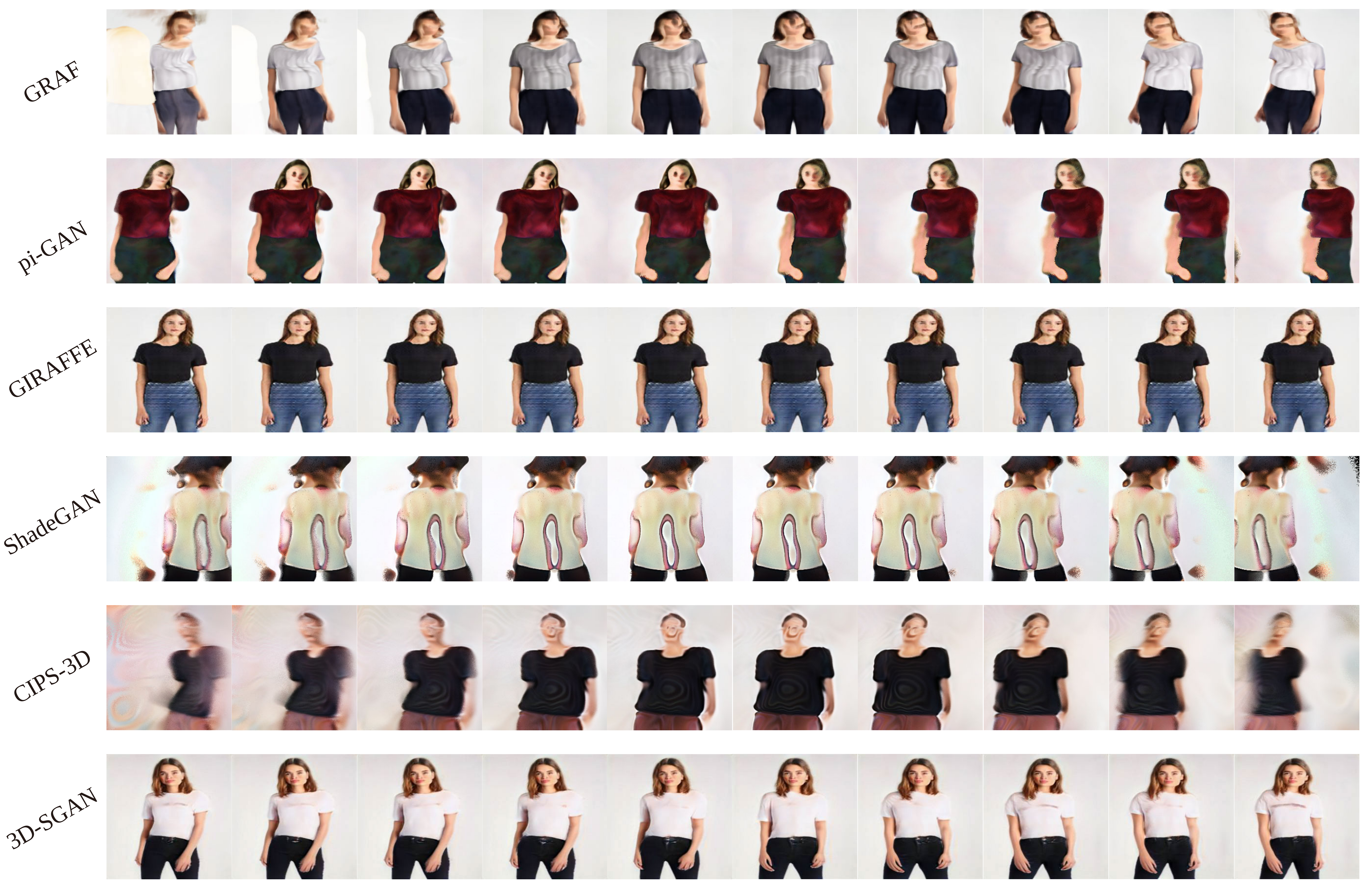}
\end{center}
\caption{VITON dataset: controllable image synthesis by  ``Rotation'' interpolation.}
\label{fig:sm3}
\end{figure*}

\begin{figure*}[t]\small
\begin{center}
\includegraphics[width=1.0\linewidth]{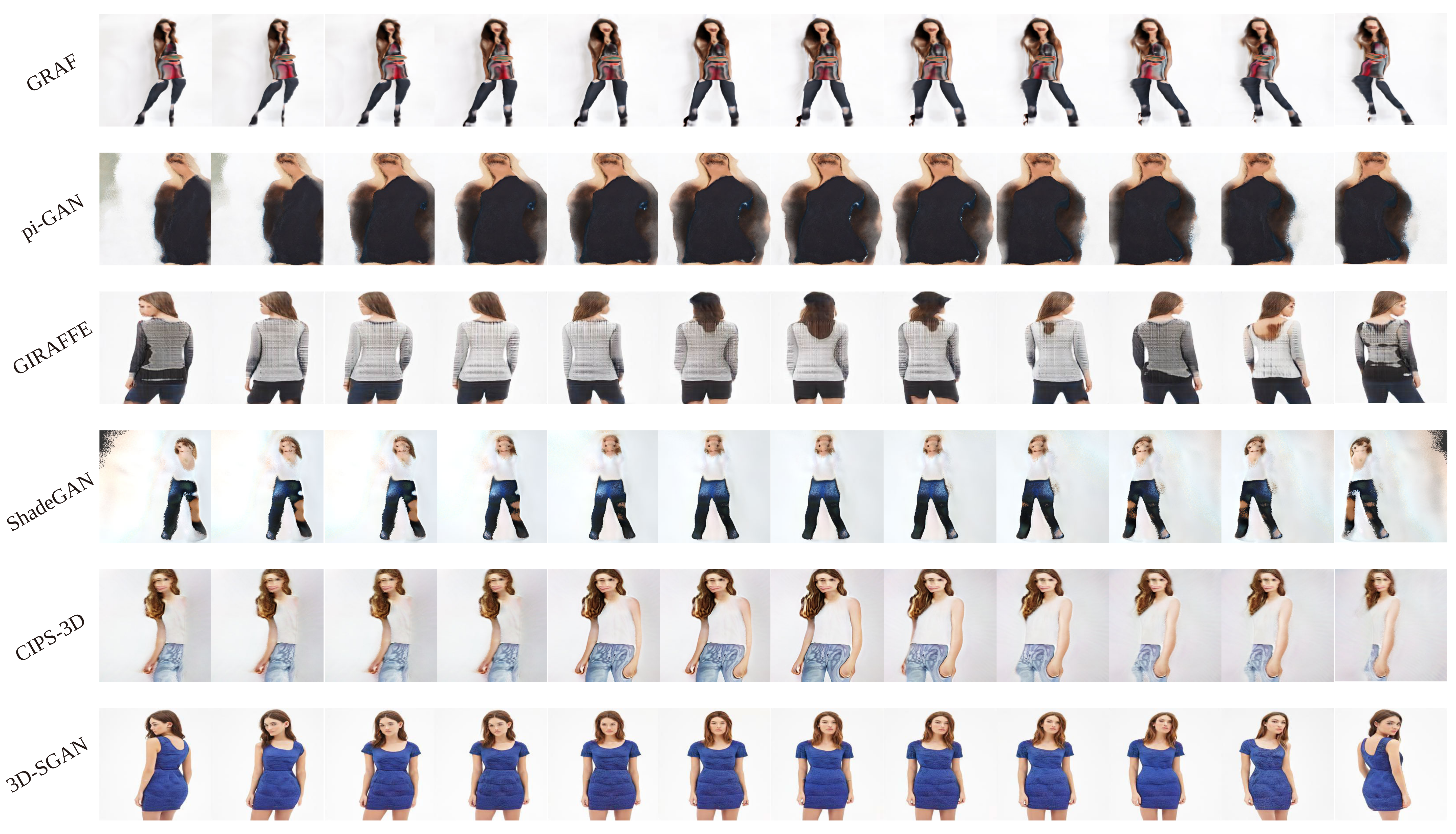}
\end{center}
\caption{DeepFashion dataset: controllable image synthesis by  ``Rotation'' interpolation.}
\label{fig:sm4}
\end{figure*}

\begin{figure*}[t]\small
\begin{center}
\includegraphics[width=1.0\linewidth]{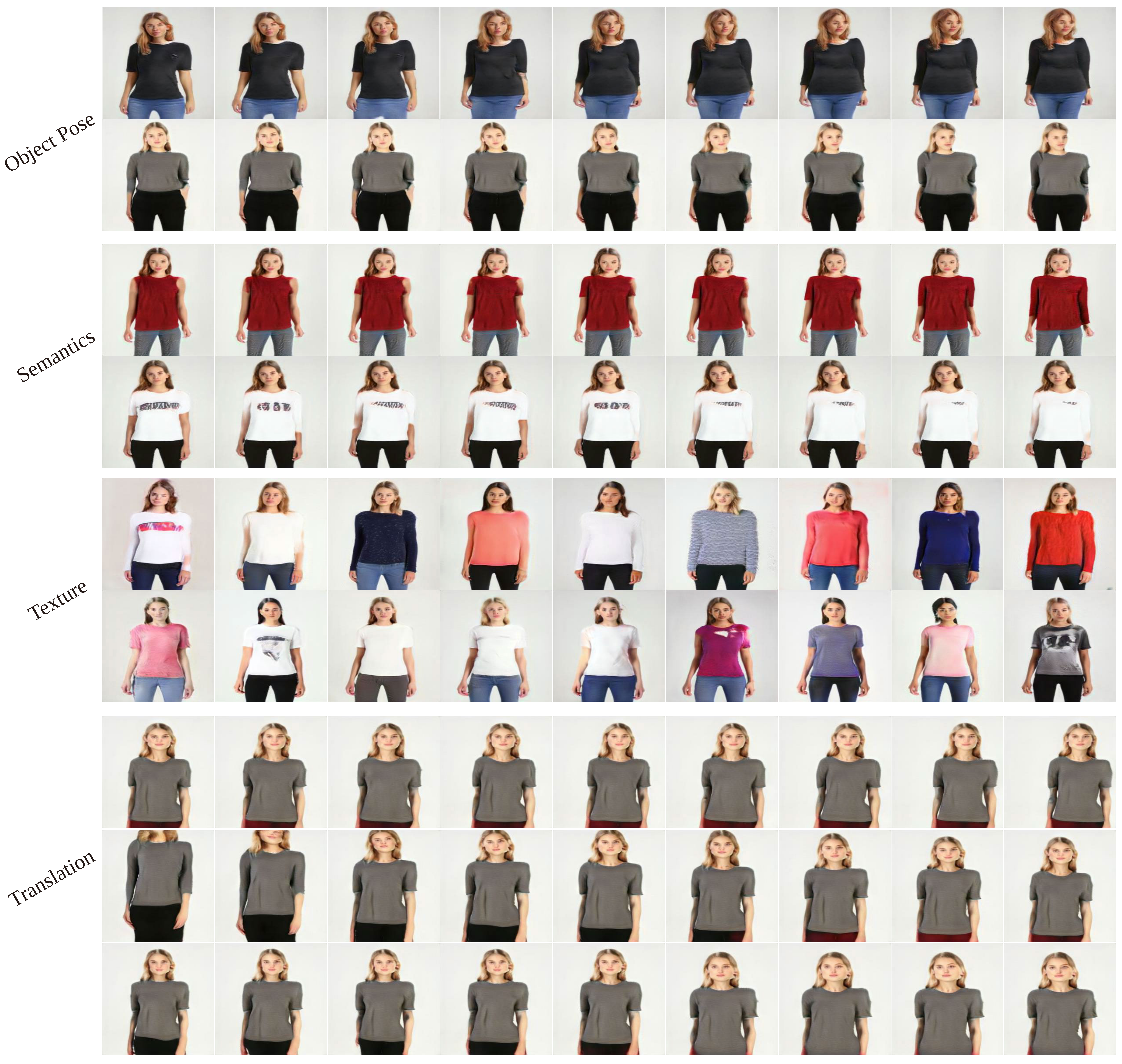}
\end{center}
\caption{VITON dataset: controllable person generation by interpolating different latent codes: `Object Pose', `Semantics', `Texture', `Translation'. For `Translation', we show generation results for `Horizontal Translation', `Vertical Translation', and `Depth Translation', respectively.}
\label{fig:sm5}
\end{figure*}

\begin{figure*}[t]\small
\begin{center}
\includegraphics[width=1.0\linewidth]{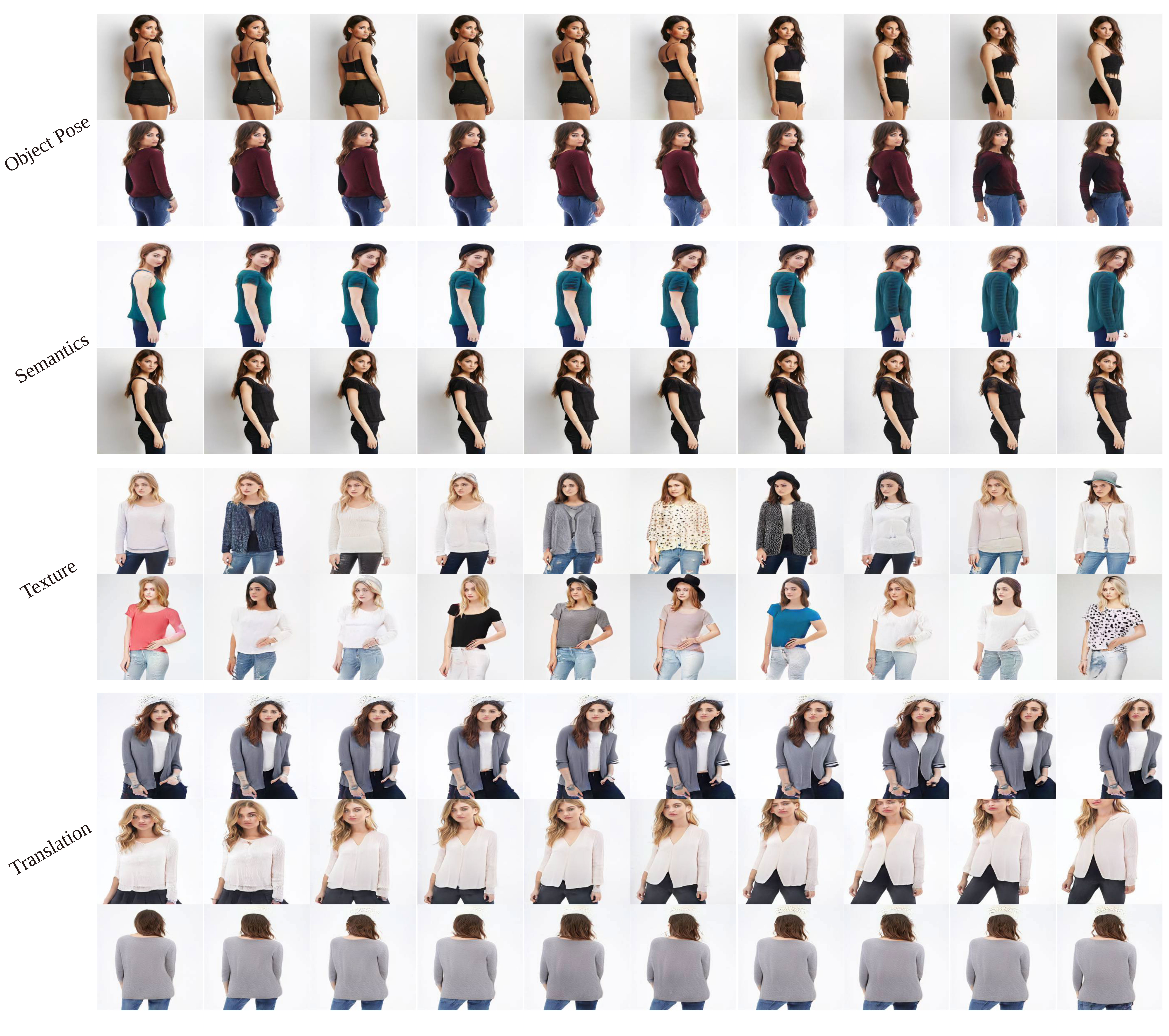}
\end{center}
\caption{DeepFashion dataset: controllable person generation by interpolating different latent codes: `Object Pose', `Semantics', `Texture', `Translation'. For `Translation', we show generation results for `Horizontal Translation', `Vertical Translation', and `Depth Translation', respectively.}
\label{fig:sm6}
\end{figure*}

\textbf{Controllable human generation.}
Fig.~\ref{fig:sm3} and Fig.~\ref{fig:sm4}
 show a qualitative  comparison using  controllable human generation where we  interpolate the ``Rotation'' parameter. Our method generates  more realistic and more view-consistent results than the baselines. Moreover, Fig.~\ref{fig:sm3} shows that GIRAFFE struggles to generate human images with different viewpoints. This is probably due to the fact that the degree for the field of view is small, and GIRAFFE may neglect the camera code and make the pose code learn most of the variations of the human body. 
Additional  controllable human generation results are shown in Fig.~\ref{fig:sm5} and Fig.~\ref{fig:sm6}. 

\begin{figure*}[t]\small
\begin{center}
\includegraphics[width=1.0\linewidth]{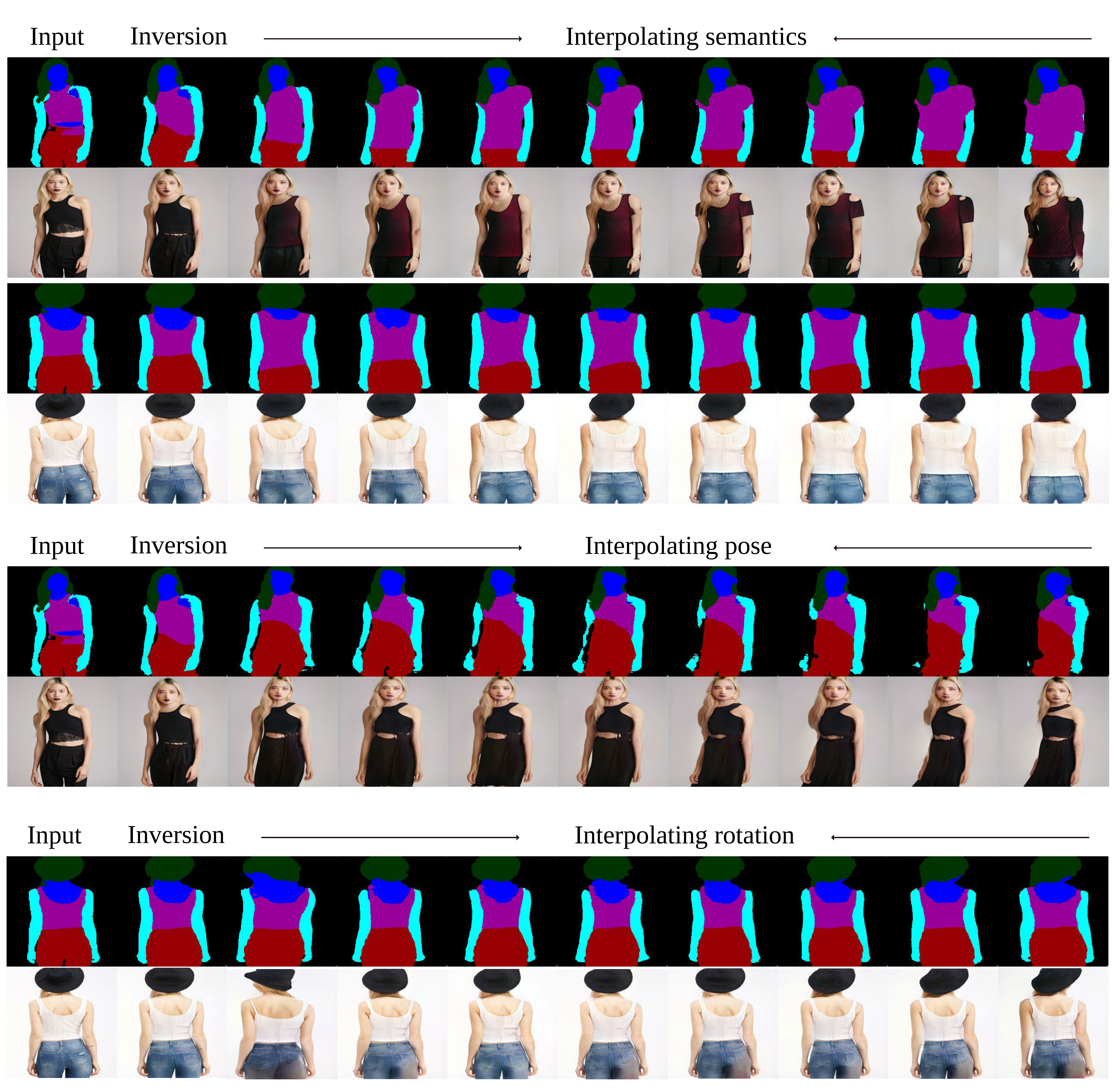}
\end{center}
\caption{DeepFashion testing dataset: real data reconstruction and editing.}
\label{fig:sm7}
\end{figure*}

 \textbf{Real human image editing.}
 Fig.~\ref{fig:sm7} shows real human image reconstruction and editing results (see the main paper for the methodological details). Our model can reconstruct the real data with only minor differences with respect to the reference image (Fig.~\ref{fig:sm7},  Column 2). The other columns of Fig.~\ref{fig:sm7} show the results of our method using latent code interpolations. Specifically, 
in top block of rows, we show   semantics editing, in the second block, pose interpolation, and in last block, rotation editing.
All the interpolations
 preserve the identity of the generated persons. 
 
\textbf{Visualization of the learning geometry.}

\begin{figure*}[t]
\begin{center}
\includegraphics[width=1.0\linewidth]{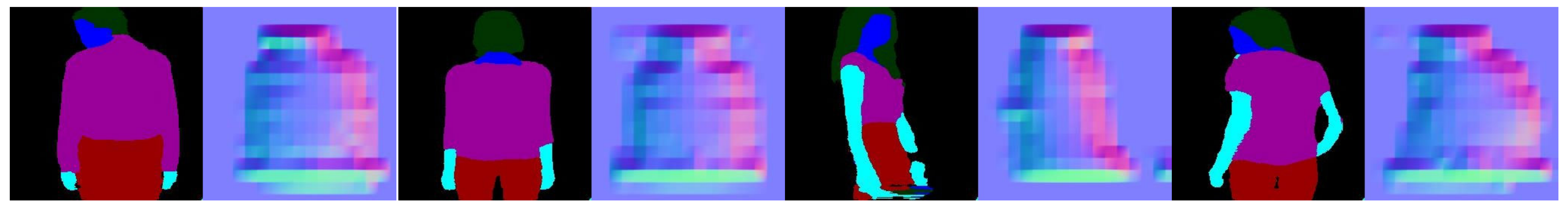}
\end{center}
\caption{The visualization of 3D features using the normals.}
\label{fig:sm_normal}
\end{figure*}

Our 3D generator is similar to GIRAFFE, from which  we  adopted the low resolution feature rendering  at $16 \times 16$, thus also our  3D predictions are not very informative. 
As shown in Fig.~\ref{fig:sm_normal}, we render the coarse normals from the density values.

\subsection{Discussion}

\subsubsection{Separate training vs. joint training}
\label{sec:SeparateTraining}

As described in the main paper, $G_{3D}$ and $G_t$ are trained separately using the segmentation tensors as a bridge between the two generators. Note that a different solution, in which 
 $G_{3D}$ and $G_t$ are jointly trained e.g., using adversarial learning, is  possible. However, in our preliminary  results, this solution led to significantly lower FID scores. Moreover,  a solution in which the same generator is in charge of modeling both the 3D structure and the texture of the data, is  conceptually similar  to GIRAFFE, whose results are significantly inferior to our proposal in this full-body generation task. We presume the reason is that both DeepFashion and VITON are relatively small and GIRAFFE struggles to learn all the variation factors, including the human pose distribution, etc. (see Sec. 1 of the main paper). In contrast, 3D-SGAN splits the problem and the architectural design in two stages. The advantages are that, this way, we can: (1) simplify the learning problem, (2) use ground truth segmentation masks (automatically obtained using \cite{Badrinarayanan2017SegNetAD}) as an additional supervision (e.g., used in $\ell_r$).
 On the other hand, 
 a potential disadvantage in separate training is  a possible domain gap  between training and inference, since 
$G_t$, at inference time, is fed with segmentation tensors generated by  $G_{3D}$ which have not been observed at training time. Despite that, our empirical results show that $G_t$ is robust enough to  this domain shift.

\subsubsection{View-inconsistency}

The  largest limitation of our method is the lack of a full   3D consistency of the generated textures with respect to multiple views of the same person.  However,  this problem is shared by  most 3D-aware GANs, because training does not include  multiple {\em paired}
views of the same person or other similar supervision. Moreover, we inherit from GIRAFFE (whose structure is adopted in our $G_{3D}$) the ``mirror symmetry'' problem \cite{zhou2021CIPS3D}, which depends on  
the way point coordinates are represented ($\gamma(\pmb{x})$). Nevertheless, 
3D-SGAN can alleviate the consistency issues to a large extent. 
 Specifically, when we generate multiple views of the same person, we change the camera pose $\pmb{z}_{c}$ {\em keeping fixed} all the other latent codes. In particular, $\pmb{z}_{t}$ is {\em fixed} and it is 
 sampled {\em only once} when, e.g., we produce the  ``Rotation'' results. This way we can generate multiview images of the same person with 
 {\em the same} overall texture (e.g., for the clothes). 
 Moreover, 
 3D consistency is further encouraged by the proposed consistency losses. 
 Although we do not fully solve the problem (e.g., we cannot control the face details), nevertheless we can alleviate it, 
 as empirically shown by  the comparison with respect to the baselines.

\clearpage

%
%

\end{document}